\pdfoutput=1

\documentclass[11pt]{article}

\usepackage[final]{acl}

\usepackage{times}
\usepackage{latexsym}
\usepackage{amsmath}
\usepackage[T1]{fontenc}

\usepackage[utf8]{inputenc}

\usepackage{microtype}

\usepackage{inconsolata}

\usepackage{graphicx}
\usepackage{array}
\usepackage{tabularx}
\usepackage{longtable}
\usepackage{booktabs}
\usepackage{colortbl}
\usepackage{xcolor}
\usepackage{makecell}
\usepackage{multirow}

\newcommand{\dataset}{\textsc{M\textsuperscript{3}C}}

\usepackage{pifont}
\newcommand{\cmark}{\ding{51}}%
\newcommand{\xmark}{\ding{55}}%

\newcommand{\grayrow}{\rowcolor[gray]{0.9}}
\title{Enabling Chatbots with Eyes and Ears:\\
An Immersive Multimodal Conversation System for Dynamic Interactions}

\author{%
 Jihyoung Jang$^{1}$\thanks{~~Equal contribution.}\quad Minwook Bae$^{3*}$\quad Minji Kim$^{1}$\quad Dilek Hakkani-Tür$^{4}$\quad Hyounghun Kim$^{1,2}$\\
$^{1}$Graduate School of Artificial Intelligence, POSTECH\\
$^{2}$Department of Computer Science and Engineering, POSTECH\\
$^{3}$Artificial Intelligence Graduate School, UNIST\\
$^{4}$University of Illinois Urbana-Champaign\\
\texttt{\{jihyoung, mzkim, h.kim\}@postech.ac.kr}\quad \texttt{minwook09@unist.ac.kr}\quad \texttt{dilek@illinois.edu}\\
}

\begin{document}
\maketitle

\begin{abstract}
As chatbots continue to evolve toward human-like, real-world, interactions, multimodality remains an active area of research and exploration. So far, efforts to integrate multimodality into chatbots have primarily focused on image-centric tasks, such as visual dialogue and image-based instructions, placing emphasis on the ``eyes'' of human perception while neglecting the ``ears'', namely auditory aspects. Moreover, these studies often center around static interactions that focus on discussing the modality rather than naturally incorporating it into the conversation, which limits the richness of simultaneous, dynamic engagement. Furthermore, while multimodality has been explored in multi-party and multi-session conversations, task-specific constraints have hindered its seamless integration into dynamic, natural conversations. To address these challenges, this study aims to equip chatbots with ``eyes and ears'' capable of more immersive interactions with humans.
As part of this effort, we introduce a new multimodal conversation dataset, \textbf{M}ultimodal \textbf{M}ulti-Session \textbf{M}ulti-Party \textbf{C}onversation (\dataset{}), and propose a novel multimodal conversation model featuring multimodal memory retrieval. Our model, trained on the \dataset{}, demonstrates the ability to seamlessly engage in long-term conversations with multiple speakers in complex, real-world-like settings, effectively processing visual and auditory inputs to understand and respond appropriately. Human evaluations highlight the model’s strong performance in maintaining coherent and dynamic interactions, demonstrating its potential for advanced multimodal conversational agents.\footnote{Our code, dataset, and model are publicly available at \url{https://m3c-dataset.github.io/}.}
\end{abstract}

\section{Introduction}

\begin{figure*}[t]
    \centering
    \includegraphics[width=1.99\columnwidth]{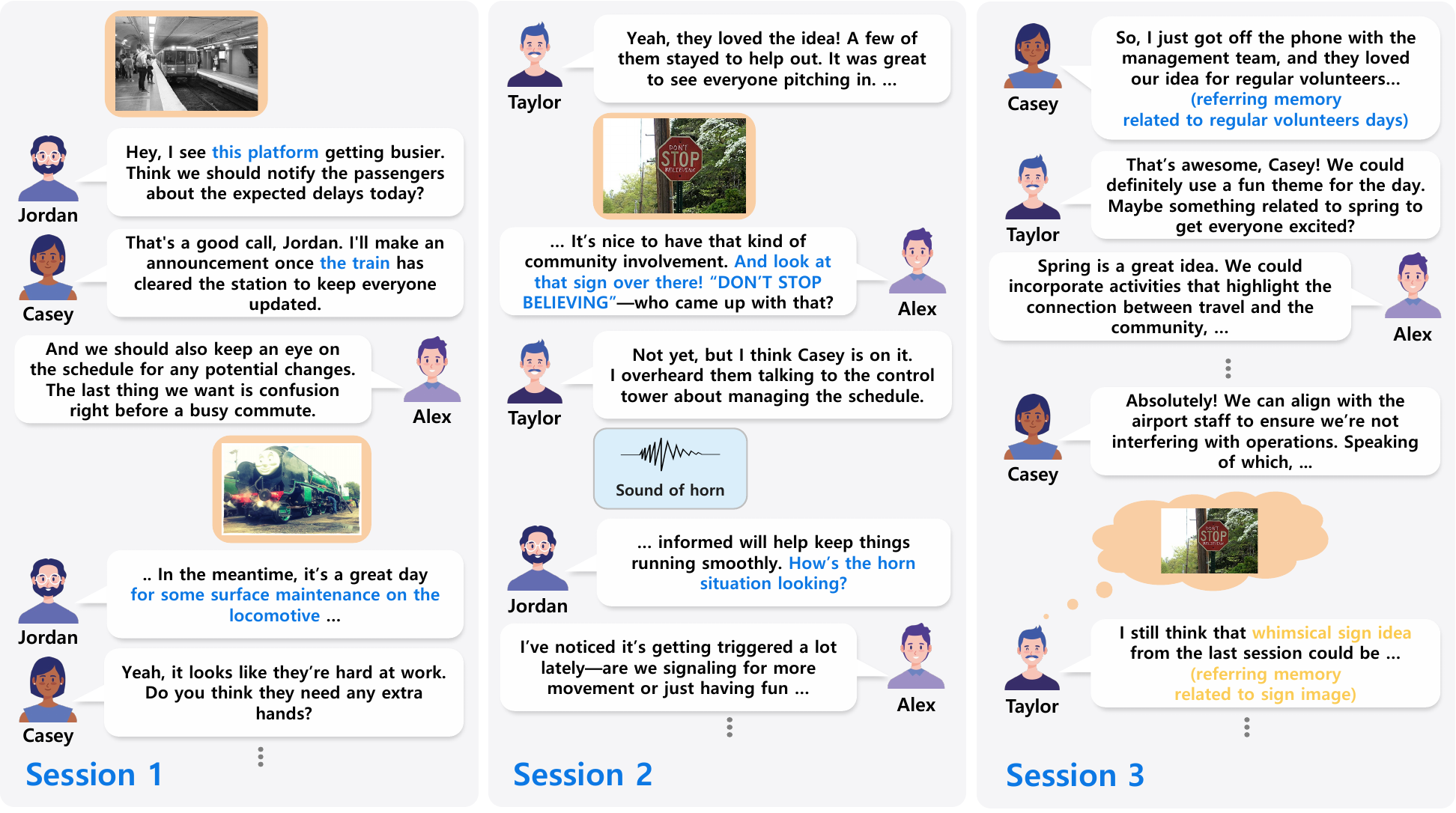}
    \caption{The main speaker (Alex) engages in conversation with two different partners per session, where all speakers simultaneously experience the provided multimodal inputs in the same shared space. In later sessions, the main speaker can meet new partners and continue the conversation.}
    \label{fig:dataset}
    \vspace{-10pt}
\end{figure*}

The development of conversation systems has made significant strides in recent years, with conversation models transitioning from simple rule-based systems to sophisticated models capable of engaging in human-like interactions~\cite{li2017dailydialog, zhang-etal-2018-personalizing, rashkin-etal-2019-towards, adiwardana2020towards, roller-etal-2021-recipes, shuster2022blenderbot, jang-etal-2023-conversation}. As these systems advance, integrating multimodal capabilities has emerged as a critical avenue for enhancing their realism. Multimodal conversation models, which combine information from multiple sensory modalities such as text, vision, and audio, hold the promise of emulating human communication more effectively by incorporating richer contextual understanding~\cite{zhang2019generative, ahn-etal-2023-mpchat, lee-etal-2024-stark, park-etal-2024-lets}. Despite this potential, existing research has primarily focused on image-based conversations, such as visual dialogue~\cite{shuster-etal-2020-image, meng2020openvidial, wang2021openvidial,feng-etal-2023-mmdialog} and image-based instructions~\cite{li2023blip, brooks2023instructpix2pix, liu2024visual, koh2024generating, yang2024imagebrush}, emphasizing the visual, or ``eyes'', aspect of human perception.
Meanwhile, the auditory, or ``ears'', aspect remains relatively underexplored, limiting the holistic nature of current multimodal systems~\cite{goel2024audio, kong2024audio, huang2024audiogpt, gong2024listen, tang2024salmonn}. However, existing approaches have been rather limited in integrating both ``eyes and ears'' simultaneously, highlighting a gap in developing truly holistic multimodal conversational agents.

Also, a key limitation of existing approaches lies in their focus on static interactions.
In these paradigms, the chatbot typically receives a shared image and responds to questions or prompts about it, similar to exchanging observations or answers about the image rather than experiencing it as part of a shared scene~\cite{Das_2017_CVPR, kottur-etal-2021-simmc}. This approach fails to capture the dynamic, real-time nature of human communication.
Furthermore, while multimodal integration has been explored in both multi-party and multi-session conversations~\cite{10.1007/11677482_3, yamamoto2015multimodal, saha2018towards, lee2023multimodal, lee-etal-2024-stark}, existing studies are largely task-focused, with little exploration of more complex, dynamic real-world scenarios. These gaps highlight the need for datasets and models that reflect the complexities of natural human conversation, encompassing multiple modalities and interaction dynamics.

To address these challenges, we propose a novel approach for equipping chatbots with ``eyes and ears'' capable of real-time, multimodal interactions. Central to this effort is the introduction of \textbf{M}ultimodal \textbf{M}ulti-Session \textbf{M}ulti-Party \textbf{C}onversation (\dataset{}), a new multimodal conversation dataset designed to capture dynamic and rich conversational settings (Figure~\ref{fig:dataset}).
Our \dataset{} features consecutive sessions involving multiple speakers, with a main speaker interacting with multiple partners in each session.
Specifically, extending the mixed-session conversation~\cite{jang-etal-2024-mixed}, \dataset{} allows the main speaker to interact with multiple partners per session while engaging with different partners across sessions, enhancing interaction diversity.
Crucially, all speakers in each session observe and interact through simultaneous visual and auditory inputs, fostering a more cohesive and realistic interaction dynamic.

Building on this dataset, we also propose a novel multimodal conversation model.
Our model consists of a dialogue module and a retriever module, enabling it to not only process multimodal inputs from the ongoing session but also store and retrieve multimodal information from previous sessions as part of a multimodal memory.
This design seamlessly integrates visual and auditory inputs, enabling coherent and contextually relevant responses in dynamic, real-world-like conversational settings. {In human evaluations, our model demonstrates high engagement and immersion with multimodality during conversations.

Here are our contributions:

\begin{enumerate}
\itemsep0em 
\item We introduce \dataset{}, a new multimodal conversation dataset featuring multiple speakers simultaneously experiencing the same visual and auditory inputs in a shared spatial and temporal context across consecutive sessions.

\item We present a novel multimodal conversation model leveraging a multimodal memory retriever, enabling it to recall and retrieve past visual and auditory elements for contextually rich and coherent responses across sessions.

\item In human evaluations, our model demonstrated deep immersion in multimodal interactions during conversations, resulting in high engagement.

\end{enumerate}
\section{Related Works}
\paragraph{Multimodal Conversation.} 
Recent work has explored integrating multiple modalities to enhance conversational interactions~\cite{Das_2017_CVPR, poria-etal-2019-meld, saha2018towards, kottur-etal-2021-simmc, lee-etal-2024-stark}, but most studies emphasize the visual aspect, effectively giving chatbots ``eyes'' while neglecting ``ears''. As a result, multimodal conversations often remain image-centric~\cite{shuster-etal-2020-image, zang-etal-2021-photochat, lee-etal-2021-constructing, ahn-etal-2023-mpchat}, failing to capture the fluid and dynamic nature of real-world conversations. While some research addresses auditory information, it is typically limited to specific sounds or narrow domains~\cite{goel2024audio, kong2024audio, huang2024audiogpt, gong2024listen, tang2024salmonn}, restricting the scope of audio integration. Moreover, existing methods that incorporate dialogue memory and retrieval often focus on single-modality inputs (e.g., text-based summaries), thereby losing critical multimodal context~\cite{li-etal-2023-pace, zhang-etal-2024-stickerconv}. To address these limitations, we propose a new multimodal conversation dataset and model that jointly processes both visual and auditory stimuli while maintaining a broader contextual memory from images, audio, and conversation history.

\paragraph{Complex Conversation Scenario.} 
Previous studies on multi-session conversations mainly concentrate on preserving context in extended one-on-one settings~\cite{bae-etal-2022-keep, jang-etal-2023-conversation, zhang-etal-2023-mind, joko2024doing}, whereas multi-party conversations often focus on coordinating multiple speakers in a single session~\cite{wei2023multi, chen-etal-2023-places, gu-etal-2023-madnet, ma2023enhanced, fan-etal-2024-improving}.~\citet{jang-etal-2024-mixed} bridges these by allowing a main speaker to interact with different partners across sessions. Extending this idea, we introduce multi-party interactions within each session, enabling several speakers to engage simultaneously. This design better reflects the dynamic interplay of real-world conversations. Furthermore, by integrating visual and auditory inputs in every session, we ensure all participants share synchronized stimuli, promoting more cohesive and realistic interaction dynamics.
\section{\dataset{}}

We introduce a new machine-generated multimodal conversation dataset, \dataset{}. 
Each episode in \dataset{} consists of four speakers engaging in conversations across three consecutive sessions. Each session is multi-party, with a main speaker interacting with two different partner combinations per session. All speakers share the same spatial and temporal environment, experiencing two instances of multimodal input (visual or auditory) per session as they converse. Since previous machine-generated datasets have demonstrated sophistication and high quality, we follow the same approach by generating our dataset using large language models (LLMs)~\cite{kim-etal-2022-prosocialdialog, zheng-etal-2023-augesc, kim-etal-2023-soda, jang-etal-2023-conversation, xu-etal-2023-baize, lee-etal-2024-dialogcc, jang-etal-2024-mixed, lee-etal-2024-stark}. We employ GPT-4o mini\footnote{\url{https://openai.com/index/gpt-4o-mini-advancing-cost-efficient-intelligence/}} to generate the dataset.

\begin{table*}[t]
\centering
\resizebox{0.99\linewidth}{!}{
\begin{tabular}{c c c c c >{\centering\arraybackslash}p{1cm} c >{\centering\arraybackslash}p{1cm} c c}
\hline
\textbf{Datasets} & \textbf{Type} & \textbf{\makecell{Multiple\\Sessions}} & \textbf{\makecell{Multiple\\Speakers}} & \multicolumn{2}{c}{\textbf{\makecell{Image \\ (\# of Images)}}} & \multicolumn{2}{c}{\textbf{\makecell{Audio \\ (\# of Audios)}}} & \textbf{\# of Sessions} & \textbf{\# of Turns}\\
\hline
AMI~\cite{10.1007/11677482_3} & Open-Domain & \textcolor{orange}{\xmark} & \textcolor{blue}{\cmark} & \textcolor{orange}{\xmark}&  & \textcolor{blue}{\cmark} & - & 279 & - \\
VisDial~\cite{Das_2017_CVPR} & Modality-QA & \textcolor{orange}{\xmark} & \textcolor{orange}{\xmark} & \textcolor{blue}{\cmark} & {\small (120K)} & \textcolor{orange}{\xmark}&  & 123K & 2.4M\\
MELD~\cite{poria-etal-2019-meld} & Open-Domain & \textcolor{orange}{\xmark} & \textcolor{blue}{\cmark} & \textcolor{blue}{\cmark} & - & \textcolor{blue}{\cmark} & - & 1.4K & 13K \\
ImageChat~\cite{shuster-etal-2020-image} & Modality-Centric & \textcolor{orange}{\xmark} & \textcolor{orange}{\xmark} & \textcolor{blue}{\cmark} & {\small (202K)} & \textcolor{orange}{\xmark}&  & 202K & 401K\\
MMConv~\cite{liao2021mmconv} & Modality-Centric & \textcolor{orange}{\xmark} & \textcolor{orange}{\xmark} & \textcolor{blue}{\cmark} & {\small (114K)} & \textcolor{orange}{\xmark}&  & 5.1K & 39.8K\\
PhotoChat~\cite{zang-etal-2021-photochat} & Open-Domain & \textcolor{orange}{\xmark} & \textcolor{orange}{\xmark} & \textcolor{blue}{\cmark} & {\small (10.9K)} & \textcolor{orange}{\xmark}&  & 12K & 156K\\
MMDD~\cite{lee-etal-2021-constructing} & Modality-Centric & \textcolor{orange}{\xmark} & \textcolor{orange}{\xmark} & \textcolor{blue}{\cmark} & {\small (13K)} & \textcolor{orange}{\xmark}&  & 17K & - \\
MMDialog~\cite{feng-etal-2023-mmdialog} & Modality-Centric & \textcolor{orange}{\xmark} & \textcolor{orange}{\xmark} & \textcolor{blue}{\cmark} & {\small (1.53M)} & \textcolor{orange}{\xmark}&  & 1.08M & 4.92M\\
MPCHAT~\cite{ahn-etal-2023-mpchat} & Modality-Centric & \textcolor{orange}{\xmark} & \textcolor{orange}{\xmark} & \textcolor{blue}{\cmark} & {\small (153K)} & \textcolor{orange}{\xmark}&  & 15K & 42.5K\\
Audio Dialogues~\cite{goel2024audio} & Modality-QA & \textcolor{orange}{\xmark} & \textcolor{orange}{\xmark} & \textcolor{orange}{\xmark}& & \textcolor{blue}{\cmark} & - & 163K & -\\
MiSC~\cite{jang-etal-2024-mixed} & Open-Domain & \textcolor{blue}{\cmark} & \textcolor{blue}{\cmark} & \textcolor{orange}{\xmark} &  & \textcolor{orange}{\xmark}&  & 51K & - \\
DialogCC~\cite{lee-etal-2024-dialogcc} & Open-Domain & \textcolor{orange}{\xmark} & \textcolor{orange}{\xmark} & \textcolor{blue}{\cmark} & {\small (129.8K)} & \textcolor{orange}{\xmark}&  & 83K & -\\
LOCOMO~\cite{maharana2024evaluating} & Open-Domain & \textcolor{blue}{\cmark} & \textcolor{orange}{\xmark} & \textcolor{blue}{\cmark} & {\small (2K)} & \textcolor{orange}{\xmark}&  & 1.7K & - \\
Stark~\cite{lee-etal-2024-stark} & Open-Domain & \textcolor{blue}{\cmark} & \textcolor{orange}{\xmark} & \textcolor{blue}{\cmark} & {\small (900K)} & \textcolor{orange}{\xmark}&  & 500K & - \\
\hline
\textbf{\dataset{}(Ours)} & Open-Domain & \textcolor{blue}{\cmark} & \textcolor{blue}{\cmark} & \textcolor{blue}{\cmark} & {\small (24K)} & \textcolor{blue}{\cmark} & {\small (73K)} & 16K & 2.5M\\
\hline
\end{tabular}
}
\caption{Statistics and comparison with other datasets. `Type' defines the scope of the conversation; Modality-QA focuses on question-answering, Modality-Centric revolves around discussions centered on a specific modality (such as images or audio), and Open-Domain allows for interactions on a wide range of topics. The `-' denotes unreported data. Our \dataset{} integrates both images and audio, creating dynamic, real-time, and immersive experiences.}
\label{tab:dataset}
\vspace{-10pt}
\end{table*}

\subsection{Structuring Multimodality}
In our dataset construction, we represent both images and audio through textual captions, since current state-of-the-art LLMs cannot directly process multiple raw image and audio inputs at once. We use COCO~\cite{DBLP:conf/eccv/LinMBHPRDZ14} as the seed image dataset and AudioCaps~\cite{kim-etal-2019-audiocaps} and Clotho~\cite{drossos2020clotho} as the audio datasets. We observe that captions significantly influence the quality of the generated conversations; therefore, we refine the image captions using GPT-4o mini, particularly to include details that are present in the images but missing in the original captions. 

\subsection{Preparing Scenario}
Before generating each conversation, we create a corresponding scenario that includes speaker information (such as names and relationships), session-specific partners, two multimodal inputs per session, and specified time intervals between sessions. To ensure thematic consistency, we group candidates' images and audio clips based on similarity similar modalities and provide them to LLM, which then selects the most fitting modalities while considering the speaker details and the time intervals. To group similar multimodal data, we perform K-means clustering (K=30) based on location tags assigned by GPT-4o-mini based on each caption’s content to facilitate clustering.

\subsection{Conversation and Memory Generation}
We sequentially generate conversation episodes based on the prepared scenarios. According to~\citet{jang-etal-2024-mixed}, when the main speaker meets different partners in each session, it is essential to track interactions separately. In this dataset, a memory summary is created after each session from the main speaker’s perspective, integrating experiences along with images and audio from previous sessions to form a multimodal memory. We also employ memory linking to connect related elements, ensuring that linked memories are retrieved together for an enriched conversational context. After generating the conversation and memory, we perform tagging to indicate where modality elements begin in the conversation and which utterances are associated with specific memory elements.

\begin{figure*}[t]
    \centering
    \includegraphics[width=2.0\columnwidth]{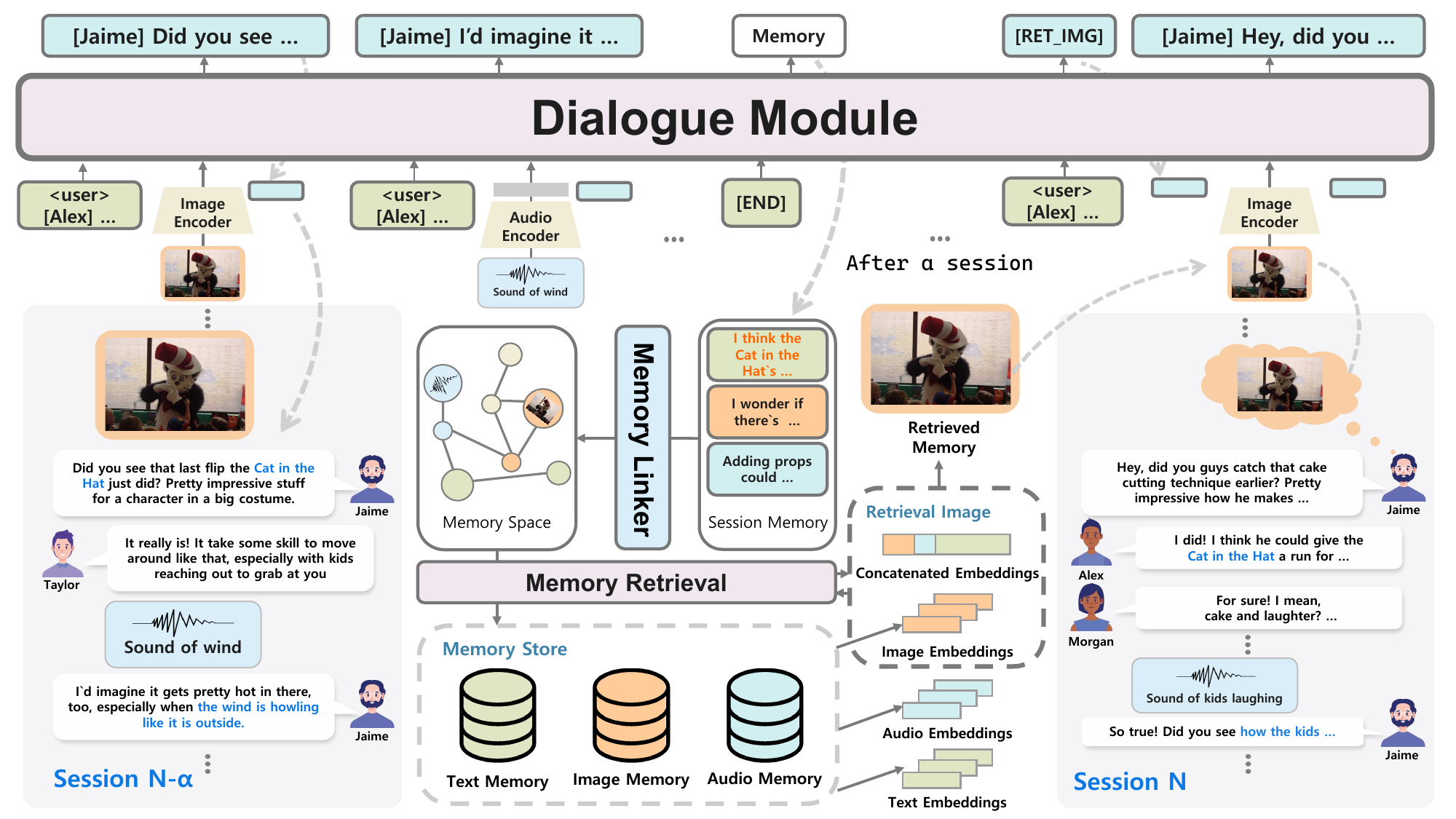}
    \caption{Overall architecture of our model. In Session N-$\alpha$, Jaime (main speaker) perceives an image of a man in a cat mask and the sound of the wind, engaging in a related conversation with Taylor. In Session N, Jaime recalls this image and linked memory to continue the dialogue. The Retrieval Module selects relevant memories—including images, audio, and chat history—based on the session context.}
    \label{fig:model}
\end{figure*}

\subsection{Filtering for Dataset Integrity}
A core requirement of our dataset is that all speakers share the same visual and auditory elements, ensuring conversations occur in a shared environment with consistent modality integration. To this end, a filtering process is employed by posing validation questions to a machine to exclude episodes that fail to maintain temporal or spatial consistency (see Appendix~\ref{sec:filtering} for more details). Through this process, we collect a total of 54K conversation episodes (34K for train, 8K for validation, and 12K for test). Table~\ref{tab:dataset} presents detailed features and statistics of \dataset{}, along with a comparison to other datasets. Also, please see Appendix~\ref{sec:prompts} for the full prompts used in building the dataset and Appendix~\ref{sec:dataexam} for examples of \dataset{} (episode and memory).
\section{Multimodal Multi-Session Multi-Party Conversation Model}
We propose a novel multimodal, multi-session, multi-party conversation model capable of perceiving both images and audio, akin to having ``eyes and ears''. It is designed to facilitate interactions among multiple partners, enabling dynamic and coherent conversations while participants change across sessions.

To achieve this, we introduce a dialogue module and a retriever module (see Figure~\ref{fig:model} for the overall architecture). The Dialogue Module performs the generation task, which includes producing conversations based on multimodal inputs, generating session memory, and linking past interactions to ensure coherence in multi-session dialogues. The Retriever Module performs the retrieval task, accessing relevant memories from previous sessions based on the ongoing session. This allows the model to integrate past memory, ensuring consistency across interactions. The following subsections detail each module.
\subsection{Dialogue Module}
The Dialogue Module performs various generation tasks, including conversation generation, memory generation, and memory linking. These processes enable the model to engage in coherent and dynamic multi-session interactions while maintaining contextual continuity. During each session, the module generates responses based on the given dialogue history and multimodal inputs, ensuring a natural and context-aware conversation flow. Once a session concludes, the model constructs a memory that captures the dialogue exchanged with different speakers and integrates perceived modalities. 

To maintain long-term coherence, the module employs structured memory linking by explicitly associating new memory units with semantically and perceptually related modalities and contextual cues. This ensures that the memory graph accurately reflects both temporal progression and meaningful relationships across sessions. For example, if the system hears the sound of wind accompanied by dialogue such as, ``Be careful, the wind might blow your hat away,'' it explicitly links the auditory perception (the sound of wind) to the semantic content of the dialogue. These associations are formed at the time of memory storage—rather than retrieval—enabling contextually relevant activations, where recalling one memory triggers related semantic or perceptual experiences. This process helps the model construct a coherent and evolving memory space. For details on the training process, please refer to Section~\ref{model:imp}.
\subsection{Retrieval Module}

The Retriever Module recalls relevant stored memories—structured through the memory linking process—based on the ongoing session’s multimodal inputs while maintaining each speaker’s perspective and sensory experiences. It retrieves past conversations, images, or audio from their respective memory stores—text from textual stores, images from visual stores, and audio from auditory stores.

To facilitate memory retrieval, the module jointly embeds the entire session—including the current conversation and the perceived modalities—into a shared representational space. It uses cosine similarity to measure how relevant the linked multimodal memories are to the present context:
\begin{equation}
sim(c, m_i) = \cos(E_c(c), E_m(m_i))
\end{equation}
where \( c \) represents the multimodal conversation context, which may include text, images, and audio, while \( m_i \) represents stored multimodal memory. \( E_c \) refers to the encoder for the conversation context, and \( E_m \) denotes the encoder for memory. During retrieval, we select only one memory with the \textit{top 1} similarity to the given context. This approach helps maintain coherence within ongoing dialogues by consistently recalling contextually relevant memories.

\section{Experiments}

\subsection{Model Training Details}
\label{model:imp}
Our model consists of a Dialogue Module and a Retrieval Module, both built on Qwen2-VL-2B-Instruct \cite{DBLP:journals/corr/abs-2409-12191}. Since the base model is a Vision-Language Model (VLM), we extend its capabilities using CLAP~\cite{DBLP:conf/icassp/WuCZHBD23}, incorporating a linear layer adapter to enable audio comprehension. The training process follows a staged tuning approach: first, we fine-tune the model on vision-language tasks, treating audio as textual captions. Then, we integrate the linear adapter, allowing the model to process audio inputs directly and enhance its auditory understanding. Also, our model supports model-to-model conversations, enabling multiple instances of the model to engage in dialogue without human intervention. To facilitate this, the model is trained to recognize and manage its own turns in a conversation, ensuring smooth turn-taking and a natural conversation flow. Further implementation details are provided in the Appendix~\ref{imp:moel_last}.

\subsection{Human Evaluation}
We conduct human evaluations to assess both the dataset quality and the model performance. For this purpose, we employ eight professional evaluators through an agency. Evaluators work in groups of three per conversation to measure inter-annotator agreement. We follow the metrics used in previous studies~\cite{lee-etal-2021-constructing, sun-etal-2022-multimodal, jang-etal-2023-conversation, lee-etal-2024-dialogcc, maharana2024evaluating, lee-etal-2024-stark, jang-etal-2024-mixed, chu2024cohesive, sumida2024should, li2024hello, lee2024thanos} for evaluation. All evaluations are based on a 5-points scale. For more details on the evaluation procedures, please refer to Appendix~\ref{sec:metrics}.
\paragraph{Dataset Quality.}
We randomly select 300 episodes from \dataset{} to evaluate overall conversation quality. We provide evaluators with instructions, the full transcripts of the conversations, and the associated modality elements. Rather than supplying captions for these modalities, we present the original image and audio files directly. Evaluators assess the conversations based on coherence and consistency, memorability, modalities alignment, and modalities engagement. Please refer to Appendix~\ref{sec:metrics} for detailed descriptions of each metric.

\paragraph{Model Performance.}
Our model is trained to enable multi-party conversations among model agents without any human intervention (please refer to Appendix~\ref{sec:modelexam} for an example). To evaluate its performance, we generate 100 conversation episodes in which only model agents participate. Each episode features four agents, and we randomly select a seed episode from our dataset to determine the agents’ names, relationships, the first utterance of the session, and multimodality content for the session.

During the conversation, each agent autonomously decides whether it is their turn to speak. If multiple agents simultaneously determine that it is their turn, the one with the higher probability takes precedence. Additionally, the main speaker decides at each turn when and which modality appears. If no modality has been introduced by the fifth turn, it is inserted at a random subsequent turn. This setup allows us to assess how effectively the model manages natural, multimodal, multi-party, multi-session conversations without human intervention. We evaluate the model’s performance on the generated episodes using Naturalness, Immersion, and Memorability, and please refer to Appendix~\ref{sec:metrics} for more detail.

\subsection{Machine Evaluation}
We conduct automatic evaluations for all conversations used in the dataset quality and model performance evaluations by employing a machine evaluator in addition to human evaluators. Specifically, we use o3-mini\footnote{\url{https://openai.com/index/openai-o3-mini/}} as the machine evaluator and provide it with the same guidelines and full conversation transcripts given to human evaluators. Unlike human evaluators, who receive direct image and audio files, o3-mini cannot accept audio inputs. Although it can process images, it cannot pinpoint the exact turn at which an image appears during the conversation. Therefore, we substitute modal content by inserting captions in the transcripts at the corresponding points.

Additionally, we also conduct comparative evaluations with other publicly available datasets, selecting those that enable fair and meaningful comparisons, specifically PhotoChat~\cite{zang-etal-2021-photochat}, DialogCC~\cite{lee-etal-2024-dialogcc}, Stark~\cite{lee-etal-2024-stark}.\footnote{Only the first sessions of Stark and \dataset{} are utilized for a fair comparison.}
For these evaluations, we randomly sample 100 conversations from each dataset and use GPT-4o-mini and Claude-3-5-Sonnet\footnote{\url{https://www.anthropic.com/news/claude-3-5-sonnet}} as machine evaluators. The models are asked to judge whether the use of multimodality in the dialogues feels natural and immersive. Evaluations are conducted in a multiple-choice format.

\begin{table}[t]
\centering
\resizebox{0.95\linewidth}{!}{
\begin{tabular}{l c c}
\hline
\textbf{Metric} & \textbf{Human} & \textbf{Machine}\\
\hline
\grayrow\multicolumn{3}{c}{\textbf{Dataset Quality}} \\
Coherence and Consistency & 4.81 & 4.99\\
Memorability & 4.63 & 4.99\\
Modalities Alignment & 4.21 & 4.26\\
Modalities Engagement & 4.36 & 4.57\\
\hline
\textbf{Dataset Overall} & 4.50 & 4.70\\
\hline
\grayrow\multicolumn{3}{c}{\textbf{Model Performance}} \\
Naturalness & 4.34 & 4.68\\
Immersive & 4.14 & 4.56\\
Memorability & 4.35 & 4.46\\
\hline
\textbf{Model Overall} & 4.28 & 4.57\\
\hline
\end{tabular}
}
\caption{Evaluation results for both dataset quality and model performance. Human score indicates the average inter-annotator rating among three groups. Machine score indicates the average evaluation score given by o3-mini.}
\label{tab:evaluation}
\vspace{-7pt}
\end{table}

\subsection{Quantitative Evaluation for Model}
We also perform automatic metric-based evaluations to measure the performace of the model.
\vspace{4pt}
\par
\noindent\textbf{Retriever Performance.}
To quantitatively evaluate the retrieval performance of our retrieval module, we assess its ability to correctly retrieve relevant memory at specific turns across all conversation episodes in the test split of the \dataset{}. 
However, since most existing baseline multimodal large language models (MLLMs) are limited to bimodal settings (such as text-image or text-audio), and models supporting all three modalities (text, image, and audio) are rare, we conduct modality-specific evaluations to ensure fair comparisons.

\paragraph{Multi-party Capability.}
Our model is designed to autonomously decide the appropriate turn to speak during multi-party conversations, which is essential for agents to interact without human involvement. To evaluate how naturally our model engages in these conversations, we randomly sample 1K episodes from the test split of our dataset. 

From each episode's first session, after the speakers have exchanged six turns of dialogue, we have the model predict which speaker would naturally respond next based on the preceding context. We compare our model's performance with a baseline (Qwen2-VL-2B-Instruct) using the same input. A prediction is considered correct if it matches the actual speaker at that turn in the dataset.
\section{Results}
\begin{table}[t]
\centering
\resizebox{0.95\linewidth}{!}{
\begin{tabular}{l c c}
\toprule
\textbf{Dataset} & \textbf{gpt-4o-mini} & \textbf{claude-3-5-sonnet}\\
\midrule
PhotoChat & 0\% & 0\%\\
DialogCC & 1\% & 0\%\\
Stark & 18\% & 1\%\\
\dataset{} & 81\% & 99\%\\
\bottomrule
\end{tabular}
}
\caption{Comparison results with other datasets using machine evaluators, showing the selection rates of each model for each dataset.}
\label{tab:judge}
\vspace{-10pt}
\end{table}
\begin{table}[t]
\centering
\resizebox{0.99\linewidth}{!}{
\begin{tabular}{l c c c}
\toprule
\grayrow\textbf{Image Only} & \textbf{R@1} & \textbf{R@5} & \textbf{MRR}\\
\midrule
Qwen2-VL-2B-Instruct~\cite{DBLP:journals/corr/abs-2409-12191} & 66.77 & 92.30 & 77.56\\
LLaVA v1.5 7B~\cite{liu2024visual} & 62.85 & 90.42 & 74.13\\
LLaMA-3.2-11B-Vision-Instruct\footnotemark & 72.41 & 90.62 & 78.90\\
Ours & \textbf{92.99} & \textbf{99.09} & \textbf{95.06}\\
\midrule
\grayrow\textbf{Audio Only} & \textbf{R@1} & \textbf{R@5} & \textbf{MRR}\\
\midrule
Qwen2-Audio-7B-Instruct~\cite{chu2024qwen2} & 69.94 & 94.85 & 80.72\\
Audio Flamingo-2~\cite{ghosh2025audio} & 63.29 & 85.44 & 70.87\\
Pengi~\cite{deshmukh2023pengi} & 68.33 & 88.75 & 75.12\\
Ours & \textbf{92.83} & \textbf{98.19} & \textbf{94.78}\\
\bottomrule
\end{tabular}
}
\caption{Experimental results on the retrieval module's performance. R@K represents Recall, and MRR indicates Mean Reciprocal Rank.}
\label{tab:retrieval}
\vspace{-5pt}
\end{table}
\footnotetext{\url{https://ai.meta.com/blog/llama-3-2-connect-2024-vision-edge-mobile-devices/}}

\subsection{Human and Machine Evaluations}
Table~\ref{tab:evaluation} presents the evaluation results for both \dataset{} and our model. The human evaluation score, derived from the average inter-annotator ratings, indicates that both the dataset and the model achieve consistently high scores across all metrics. 
Machine scores are slightly higher than human scores but show a similar distribution trend. In particular, considering the concept of images or audio appearing simultaneously to speakers, the dataset also shows high scores in modality-related metrics. For model evaluation, where four model agents engaged in conversations without human intervention, the results further demonstrate the model's ability to conduct human-like interactions. These findings confirm that our model effectively integrates visual and auditory inputs, behaving like it has ``eyes and ears'' to engage in immersive, contextually coherent conversations. As shown in Table~\ref{tab:judge}, our dataset also demonstrates higher quality compared to the other datasets, indicating that conversations in ours enable an immersive experience in which speakers naturally interact through shared spatial and temporal contexts across multiple modalities. For detailed human evaluation statistics, including average group scores and inter-annotator agreement, please refer to Appendix~\ref{sec:metrics}.
\subsection{Model Evaluations}
\paragraph{Retrieval Performance.}
Table~\ref{tab:retrieval} shows the quantitative performance of the retriever module. We use standard retrieval metrics, including Recall@K (R@1, R@5) and Mean Reciprocal Rank (MRR). As shown in the table, our model effectively identifies the relevant memory elements. Notably, it achieves a much higher R@1 score compared to other models, indicating that it’s particularly good at retrieving multimodal memory from ongoing sessions in a multi-session setting.

\begin{table}[t]
\centering
\resizebox{0.95\linewidth}{!}{
\begin{tabular}{l c}
\toprule
\textbf{Model} & \textbf{Rate}\\
\midrule
Qwen2-VL-2B-Instruct~\cite{DBLP:journals/corr/abs-2409-12191} & 10.3\%\\
Ours & 85.2\%\\
\bottomrule
\end{tabular}
}
\caption{Performance of next-speaker prediction in a multi-party setting.}
\label{tab:party}
\vspace{-5pt}
\end{table}
\begin{table}[t]
\centering
\resizebox{0.99\linewidth}{!}{
\begin{tabular}{l p{7.5cm} }
\toprule
\textbf{User} & People here look so excited! It looks so fun!\\
& \includegraphics[height=3.5cm]{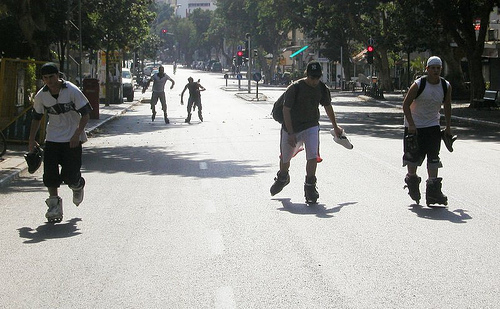} \\
\midrule
\textbf{Qwen2} & It does seem like the people in the image are enjoying themselves, as they appear to be rollerblading or inline skating on a street. ... \\
\midrule
\textbf{LLaVA} & Yes! The people in this image look like they’re having a blast rollerblading down the street. The open road, the sun, and the sense of freedom definitely add to the fun atmosphere ......\\
\midrule
\textbf{o3-mini} & It does look like they’re having a blast! From what I can see, there are several people rollerblading (or inline skating) down a street together. ......\\
\midrule
\textbf{Ours} & Yes! It’s a great day for skating. Do you think they’re just having fun, or are they trying to do a trick?\\
\bottomrule
\end{tabular}
}
\caption{A human live chat example showing the different responses from models based on image input.}
\label{tab:imgconv}  
\vspace{-7pt}
\end{table}

\paragraph{Multi-party Capability.}
The next-speaker prediction performance shown in Table~\ref{tab:party} clearly demonstrates our model’s superior ability to identify the appropriate next speaker, highlighting its deeper understanding of multi-party conversational dynamics. These results further underscore the importance of our dataset and model design in enabling more natural and coherent multi-party interactions.

\subsection{Comparison with Other Models}
Unlike existing models, our model enables all speakers to converse as if they are in the same environment, simultaneously observing and experiencing visual elements as they appear during the conversation. To validate this capability, we compare our model against its baseline, Qwen2-VL-2B-Instruct and LLaVA, both without tuning on our dataset, as well as o3-mini. We provide all models with the same context to examine how their responses differ. Table~\ref{tab:imgconv} presents the comparison examples, showing that our model integrates visual elements directly into the conversation context, responding as if the participants are physically present and viewing the image. In contrast, the other models primarily provide descriptions or fact-based responses about the image, resembling a question-and-answer interaction rather than immersive dialogue. This demonstrates that our model not only benefits from effective training on the dataset but also seamlessly adapts to the given modality, enabling more contextually rich and immersive conversations. Please see Appendix~\ref{sec:ExamMode} for additional comparison examples.

\begin{table}[t]
\centering
\resizebox{0.99\linewidth}{!}{
\begin{tabular}{l p{7.5cm} }
\toprule
\textbf{} & \includegraphics[height=3.5cm]{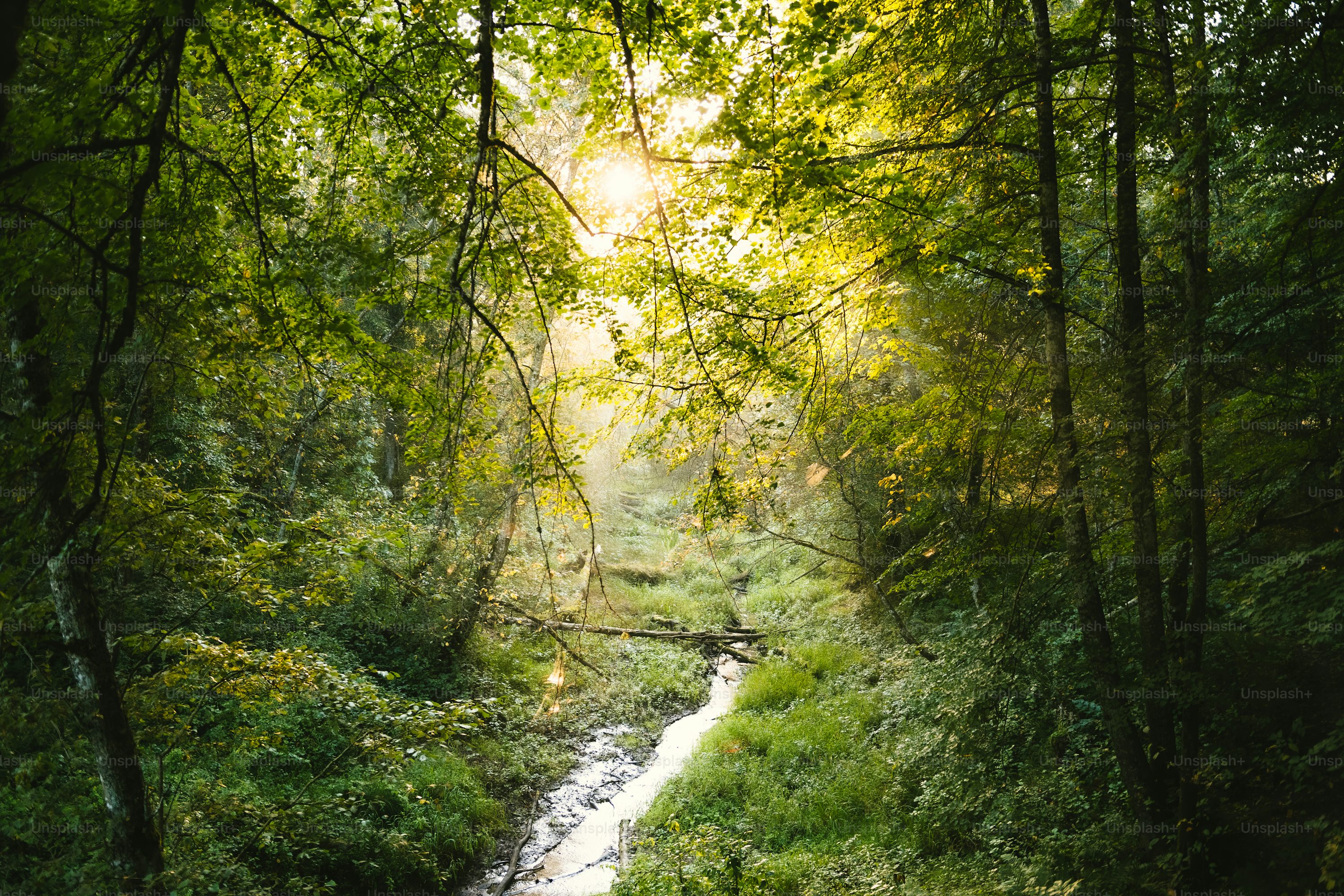}\\
\textbf{Jamie} & This place is nice. Kinda has that open, peaceful feel to it.\\
\textbf{Taylor} & Yeah, it's got that same relaxed vibe. Also, the trees here are huge!\\
\textbf{Alex} & Right? And the way the light comes through the leaves—it's pretty cool.\\
\textbf{Jamie} & This kind of reminds me of that time at the park. You know what I mean?\\
\midrule
\grayrow \textbf{Retrieved Memory} & I need to remember that we plan to target the park for our next community clean-up and set a date for this weekend (about Jamie)\\
\textbf{Alex} & Right! Maybe we can use the park for our clean-up activity. That would be great!\\
\bottomrule
\end{tabular}
}
\caption{A human live chat example showing the possibility of a long-term session than the third session in our model.}
\label{tab:longterm}  
\vspace{-10pt}
\end{table}

\subsection{More Than Three Sessions}
Our dataset consists of three consecutive sessions, and the model is trained on these three sessions. However, since our model manages the contents of previous sessions through a separate memory mechanism, it is able to maintain long-term conversations without being limited by the model’s context length. Table~\ref{tab:longterm} presents an example that demonstrates this capability. After the third session is completed and the fourth session begins, the main speaker retrieves information from the first session at an appropriate time and uses it during the conversation. This shows that although our model is trained on three sessions, it can effectively be utilized in longer sessions as well.

\begin{table}[t]
\centering
\resizebox{0.99\linewidth}{!}{
\begin{tabular}{l p{7.5cm} }
\toprule
\multicolumn{2}{l}{\bf Audio Caption vs. Raw Audio}\\
\midrule
\multicolumn{2}{l}{......}\\
\textbf{User} & It's so loud, my head hurts.\\
\midrule
\grayrow \multicolumn{2}{c}{\bf A helicopter engine runs and wind blows}\\
\grayrow \multicolumn{2}{c}{\bf Clotho dataset - 95383.wav}\\
\midrule
\textbf{Caption} & I've been thinking about the flight we had last week. It was a blast, but I'm curious about the weather conditions.\\
\midrule
\textbf{Raw Audio} & Yeah, right? I don't think we've seen anything like this before. Did we make plans after our flight earlier to hang around?\\
\bottomrule
\end{tabular}
}
\caption{A human live chat example during a helicopter ride, illustrating an ablation example of the audio adapter.}
\label{tab:AudioAbla}  
\end{table}

\subsection{Ablation Study}
\paragraph{Audio Caption vs. Raw Audio.} Our model enhances audio comprehension by integrating an audio encoder. To demonstrate the effectiveness of this audio adapter, we conduct ablation experiments comparing the model's performance on audio inputs with and without the adapter. In the model without the adapter, audio sources are provided as textual captions, while the adapted model processes raw audio clips directly. Table~\ref{tab:AudioAbla} illustrates this comparison, showing that the caption-based model heavily relies on the provided caption, often generating responses that reflect the caption’s content. For example, the caption refers to a previous flight and weather conditions, which may not be contextually linked to the actual audio. In contrast, our adapter-equipped model generates responses that align more closely with the auditory experience, such as acknowledging the loudness of the helicopter engine, demonstrating that the adapter enables the model to perceive and respond to audio more naturally and accurately.

\begin{table}[t]
\centering
\resizebox{0.99\linewidth}{!}{
\begin{tabular}{l p{7.5cm} }
\toprule
\grayrow & \includegraphics[height=3.5cm]{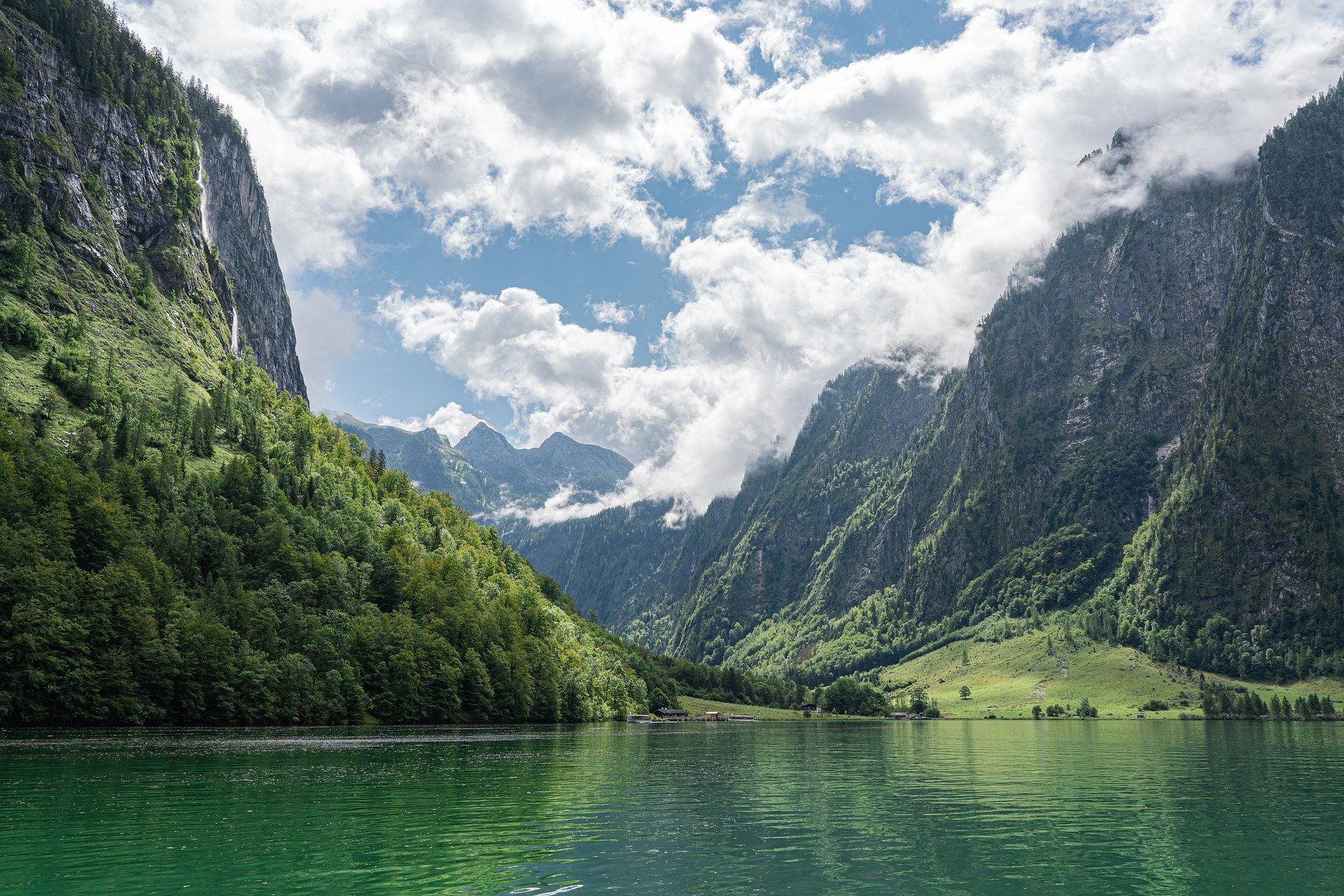} \\
\textbf{Jamie} & Wow, the view here is incredible! We have to take some photos. Who wants to go first?\\
\textbf{Alex} & I'm up for it! I've got my camera with me. Just let me know when we're ready.\\
\textbf{Morgan} & I think we have enough room! I'll just grab my camera, too.\\
\textbf{Alex} & Perfect! Let's find the best spot for the shot. Maybe over by that railing?\\
\textbf{Jamie} & This reminds me of when we were at the beach! The way the sunlight hit the water—it was so perfect.\\
\midrule
\grayrow \multicolumn{2}{c}{\bf With Retrieval}\\
\grayrow & \includegraphics[height=3.5cm]{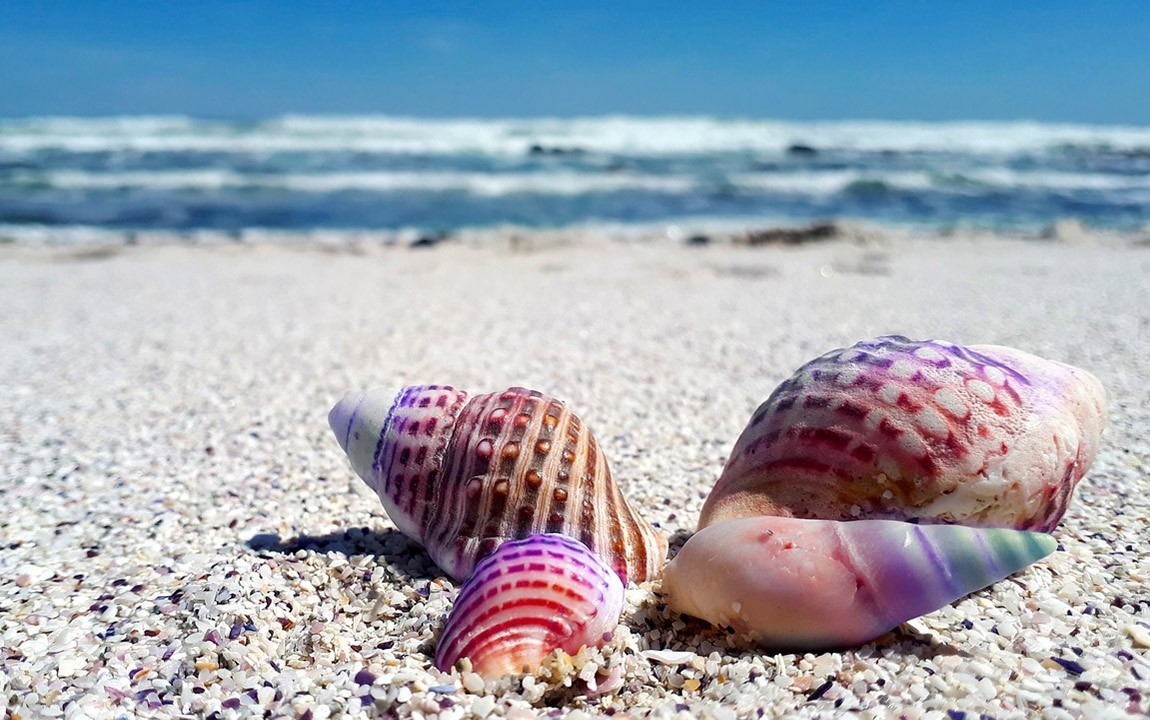} \\
\textbf{Alex} & Yeah! And it'd be fun to add some seashells to the pictures, just like those we collected the last time.\\
\grayrow \multicolumn{2}{c}{\bf Without Retrieval}\\
\textbf{Alex} & Yes! And we should avoid the crowds today. Just want to capture this view peacefully.\\
\bottomrule
\end{tabular}
}
\vspace{-5pt}
\caption{An example of a human live chat showing the difference in responses when generating utterances using a multimodal retriever with and without referencing multimodal memory.}
\label{tab:MemAbla}  
\vspace{-10pt}
\end{table}

\paragraph{Multimodal Memory.} 
Our model leverages a multimodal memory retriever to retrieve multimodal memories based on the given conversational context. To verify the effectiveness of this retriever, we compare model responses with and without the retriever under the same conversational context. Table~\ref{tab:MemAbla} illustrates this comparison, showing that when the retriever is used, the model references multimodal memories from previous session related to the beach, such as collecting seashells, enriching the conversation with relevant and vivid details. In contrast, without the retriever, the model produces a more generic response focused only on the present view. This demonstrates that our multimodal memory retriever effectively retrieves contextually relevant memories, enhancing the conversational flow and depth.
\section{Conclusion}

In this work, we address key challenges in multimodal conversation modeling. These challenges include the lack of datasets that capture dynamic and realistic interactions, the limited integration of simultaneous visual and auditory modalities, and the absence of robust models capable of handling complex conversational settings. To overcome these issues, we introduce \dataset{}, which enables immersive and authentic interactions by allowing speakers to directly observe and hear synchronized visual and auditory inputs. Building on this, we propose a novel robust multimodal conversation model designed to process and respond to both image and audio modalities concurrently. Trained on \dataset{}, our model shows strong performance in maintaining conversational coherence, adapting to diverse multimodal inputs, and managing complex interactions involving multiple partners within each session, as validated through human evaluations.

\section*{Limitations}

In building our dataset, we enhance the seed image dataset captions to improve dialogue quality. However, we do not refine the captions for the seed audio dataset. Future work that improves audio captions, similar to image captions, is expected to enhance the dataset’s audio immersion and overall quality. Additionally, our model integrates audio and image recognition by attaching an adapter to the pre-trained VLM model (Qwen), due to the lack of an instruct-tuned model capable of jointly understanding images, audio, and text. The development of an instruct-tuned Visual-Audio-Language Model (VALM) could further improve performance in future research.
\section*{Ethics Statement}

We use OpenAI’s text-moderation-latest model~\cite{markov2023holistic} to review all generated conversations for flags in categories such as sexual content, harassment, hate speech, self-harm, and violence. Any episode flagged in any of these categories is excluded from our dataset. However, we do not conduct additional ethical verification for the COCO, AudioCaps, and Clotho datasets used to provide image and audio modalities, trusting that the original authors have already performed thorough ethical reviews. For human evaluations, we commission the evaluation process to a professional agency to ensure fairness and impartiality, with no direct involvement from the authors. We also confirm that evaluators receive fair compensation and appropriate treatment through the agency. Also, our research results are to be used for research purposes only.
\section*{Acknowledgments}
We thank the reviewers for their valuable feedback. This work was supported by Institute of Information \& communications Technology Planning \& Evaluation (IITP) grant funded by the Korea government(MSIT) (No.RS-2019-II191906, Artificial Intelligence Graduate School Program(POSTECH)) and (No.RS-2020-II201336, Artificial Intelligence graduate school support(UNIST)).

\bibliography{custom}

\appendix

\section{Dataset Filtering}
\label{sec:filtering}
We verify data integrity using GPT-4o-mini with the following questions:

\begin{itemize}

\item Is there complete consistency between the environmental, spatial, and temporal features of the settings within the session? For example, it would be contradictory if one setting depicts daytime while the other depicts nighttime, or if spatial features (e.g., location or layout) and time progression are logically inconsistent. (Yes or No)

\item Do all sessions within the episode maintain a plausible continuity in time, space, and context? For example, any stated time intervals or implied transitions between settings should be logical and coherent. (Yes or No)

\item Are all participants depicted as fully engaging with the setting in real time? References to past or future events should not imply detachment from the present interaction (e.g., avoiding phrases like ``for our next scene'' or references to reviewing recorded footage). (Yes or No)

\item Are all settings within the session entirely realistic? Any elements that seem exaggerated, cartoonish, or overly stylized for natural conversation or interaction should be avoided. (Yes or No)

\item Is each setting fully utilized and referenced in the conversation? All settings presented within the session must have a clear role in the dialogue or interaction, without any being neglected. (Yes or No)

\item Do all spoken lines reflect the tone and context of natural, real-time interaction? For instance, lines should avoid referring to the setting or events in a way that suggests they are pre-recorded, staged, or viewed from an external perspective. (Yes or No)

\end{itemize}

Then, we perform basic filtering by checking that all speaker names within each episode are unique, ensuring that the two partners in each session are distinct, confirming that each conversation has at least eight turns, and verifying that all three partners participate in at least one session. Any episodes that fail to meet these criteria are excluded from the dataset.

\section{Prompts List}
\label{sec:prompts}
We use gpt-4o-mini to construct our \dataset{}. Table~\ref{tab:PromCaption} shows the prompt for caption refinement, Table~\ref{tab:PromLocation} shows the prompt for location assignment, Table~\ref{tab:PromCandidate} shows the prompt for scenario generation, Table~\ref{tab:PromAlign} shows the prompt for caption pair validation, Table~\ref{tab:PromSession} shows the prompt for session conversation generation, Table~\ref{tab:PromTagging} the prompt for shows modality tagging, Table~\ref{tab:PromMemGen} shows the prompt for memory summarization, Table~\ref{tab:PromMemLink} shows the prompt for memory linking, Table~\ref{tab:PromMemTag} shows the prompt for memory tagging, and Table~\ref{tab:PromDataValid} shows the prompts for episode validation.

\section{Dataset Examples}
\label{sec:dataexam}
Please refer to Table~\ref{tab:DataExam1}, Table~\ref{tab:DataExam2}, and Table~\ref{tab:DataExam3} for conversation examples of \dataset{} episodes, Table~\ref{tab:DataMemSum} for an example of memory summarization, and Table~\ref{tab:DataMemLink} for an example of links between multimodal memories.

\section{Implementation Details}
\label{imp:moel_last}
We use all pre-trained models through Hugging Face Transformers~\cite{wolf-etal-2020-transformers}. Our model is built based on the Qwen2-VL-2B model, using exactly 4B parameters (2B for the dialogue module and 2B for the retrieval module). Please see below for the implementation details of each module of the model.
\subsection{Dialogue Module}
The Dialogue Module is designed to process multiple modalities for generation tasks such as conversation generation, summarization, and memory linking. By leveraging the multitasking capabilities of the vision-instruct model, the module is fine-tuned with task-specific instructions, system prompts, and prefixes. This allows a single model to efficiently handle diverse tasks like dialogue generation and summarization.

\begin{table}[t]
\centering
\resizebox{0.99\linewidth}{!}{
\begin{tabular}{l p{7.5cm} }
\toprule
\multicolumn{2}{l}{\bf Example of summary generation}\\
\midrule
& ......\\
& \includegraphics[height=3.5cm]{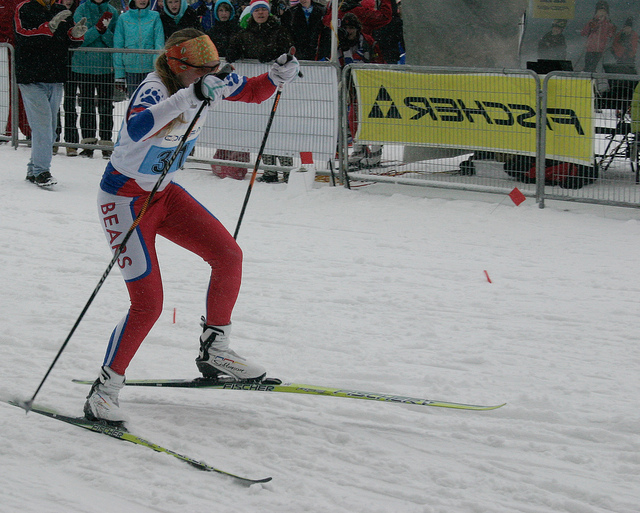} \\
\textbf{Alex} & Yeah, it takes a lot of skill to navigate through the track like that. I wonder what kind of techniques they use to keep their balance while pushing off. \\
\textbf{Morgan} & It's all about the rhythm, I guess. Just like in a game of polo, you need to know when to hit and when to pull back. \\
\textbf{} & .....\\
\textbf{User} & \textbf{[SUMMARY]}\\
\midrule
\textbf{Summary} & \textbf{I find it fascinating how much skill is required to maintain speed and balance in cross-country skiing, which seems similar to other sports.} (from first session, about myself) \textbf{<sep>} Jamie highlighted the importance of rhythm in skiing, comparing it to the game of polo, which makes me think about the coordination needed in both sports. (from first session, about Jamie) \textbf{<sep>} .....\\
\bottomrule
\end{tabular}
}
\caption{Example of memory generation across the entire session using the [SUMMARY] token. The memory includes both image-related content and conversation history, with different memory segments separated by the <sep> token.}
\label{tab:memory_exp}  
\end{table}

\paragraph{Conversation Generator.}
Our model generates responses by considering several factors, including the speaker’s identity, the ongoing conversation history, and relevant memories. It also supports model-to-model conversations without human intervention, with each model acting as a distinct speaker in a multi-party conversation. When a model wants to speak, it generates a [YES] token. If only one model generates this token, it takes the speaking turn. If multiple models generate the token, the one with the highest probability is selected to speak, ensuring smooth turn-taking and a natural conversation flow. If the model determines that introducing an image or audio is appropriate, it generates [RET\_IMG] or [RET\_AUDIO], respectively. Otherwise, it generates [NO\_RET].

\paragraph{Memory Summarizer.}
Our model summarizes the dialogue history into a personal memory at the end of each session, capturing information about itself and its conversation partners. This process takes the entire session history and a [SUMMARY] token as input, while the system prompt specifies the perspective from which the memory should be generated. When multiple memory entries are created, they are separated using the <sep> delimiter.

Table~\ref{tab:memory_exp} presents an example of memory generation using the [SUMMARY] token. This memory includes both dialogue and image-related information, with multiple segments separated by the <sep> token. In this example, Alex mentions techniques for maintaining balance in cross-country skiing, and the summary highlights how this skill is similar to other sports, such as polo.

\paragraph{Memory Linker.}
We adopt a memory-linking approach where newly generated memories are connected to relevant past ones~\cite{jang-etal-2024-mixed}. While~\citet{jang-etal-2024-mixed} propose methods for establishing memory connections within a session, extend this approach to link not only dialogues within a session but also multimodality associated with memory. This allows for the formation of a richer memory network that integrates multiple modalities rather than relying solely on text-based memories. The model determines these links by responding [POSITIVE] for relevant memories and [NEGATIVE] otherwise. 

\paragraph{Training Strategy.}
The Dialogue Module generates utterances based on audio, images, and text within a session. It also handles tasks such as summarization and memory linking. To train the model for these multi-task operations, the training process is divided into four distinct stages, as outlined below.

\begin{enumerate}
    \item \textbf{Session Utterance}\\ The model is trained to process session utterances. Images are directly provided as input without modification. However, audio is converted into captions and represented in the format \texttt{<start\_audio>} caption \texttt{<end\_audio>}. During this stage, the model learns using only the text-based representation of audio.
    
    \item \textbf{Main Speaker \& Memory Retrieval}\\ The model is trained when to output turn tokens and retrieval tokens. It decides when to take the role of the main speaker by producing a [YES] or [NO] token. Similarly, it learns the timing for memory retrieval using tokens like [RET\_IMG] for visual memory, [RET\_AUDIO] for auditory memory, and [NO\_RET] when retrieval isn't needed. 
    
    \item \textbf{Summarization \& Memory Linking}\\ 
The model is trained to generate session summaries to create structured memory. Also, it learns to establish memory links between session memory and observed images/audio, as well as between session memory and chat history. 

    \item \textbf{Audio Understanding via CLAP}\\ 
In the final stage, we enhance the model with a frozen CLAP, using a linear layer adapter to enable audio comprehension. Instead of using only captions, the model now receives audio embeddings between \texttt{<start\_audio>} and \texttt{<end\_audio>}, allowing for direct audio understanding and richer multi-modal interactions.
\end{enumerate}
Each stage progressively enhances the model’s capabilities, ensuring it can handle multimodal dialogue generation and memory management. For detailed training settings, please refer to the section below.

\paragraph{Training Setting.}
We apply the LoRA~\cite{DBLP:conf/iclr/HuSWALWWC22} to fine-tune Qwen2-VL-2B-Instruct, training with a cross-entropy loss, a maximum input length of 1024, and a learning rate of $1\times10^{-3}$. We use early stopping with a maximum of 3 epochs. For LoRA configuration, we set $r=8$, $\alpha=16$, dropout=0.05, and apply it to the q\_proj and v\_proj layers. During training stages 1, 2, and 3, we use a batch size of 32, while in stage 4, where the audio adapter is trained, we reduce the batch size to 16. Training is conducted on 8 NVIDIA RTX A6000 GPUs. We employ the following system prompts and sequence inputs for each task.

\paragraph{System Prompt for Conversation Generation}
``\texttt{$<<$SYS$>>$Please generate the speaker's next utterance. Main-Speaker: [MAIN SPEAKER NAME]-MAIN SPEAKER JOB [SUB SPEAKER NAME]-SUB SPEAKER JOB [MEMORY] MEMORY SENTENCE 1 [LINK] LINK MEMORY SENTENCE [LINK] ... [MEMORY] MEMORY SENTENCE N [LINK] $<<$$/$SYS$>>$}''

\paragraph{System Prompt for Memory Summarization}
``\texttt{$<<$SYS$>>$Please review this session's conversation history and summarize the key points [ABOUT WHO] needs for the next session. Separate each memory with <sep>. ...$<<$$/$SYS$>>$\{Full Sessions\}[SUMMARY]''}

\paragraph{System Prompt for Memory Linking}
``\texttt{$<<$SYS$>>$Please determine whether the two memory elements are related or if they reflect an update. [POSITIVE] or [NEGATIVE]$<<$$/$SYS$>>$memory sentence 1: \{MEMORY 1\} memory sentence 2: \{MEMORY 2\}''}

\subsection{Retrieval Module}
\label{imp:ret}
This module retrieves multimodal memory from prior sessions based on the context of the ongoing session, preserving speakers’ perspectives and sensory experiences by recalling past conversations, images, and audio. However, effectively retrieving such multimodal memory requires a model capable of jointly encoding diverse modalities. Traditional uni-modal encoding models, such as CLIP~\cite{DBLP:conf/icml/RadfordKHRGASAM21} or CLAP~\cite{DBLP:conf/icassp/WuCZHBD23}, are limited to processing specific modality pairs (e.g., image-text, audio-text), making it challenging to represent complex multimodal data, where image, audio, and text are freely concatenated and must be jointly encoded within the same embedding space. To overcome this limitation, inspired by~\citet{DBLP:journals/corr/abs-2410-05160}, we build a unified multimodal retrieval model that directly concatenates image, audio, and text representations within a shared embedding space, enabling efficient cross-modal memory retrieval. 
\paragraph{Training Strategy.}
As illustrated in Figure~\ref{fig:ret}, this model is trained using a contrastive learning framework that transforms a vision-audio-language model into a multi-modal embedding model, effectively aligning image, text, and audio representations. To construct a relevant query-target pair, the query consists of the current session's conversation along with perceived images or audio, represented as \( q^+ \). The target, \( t \), corresponds to a memory retrieved from previous sessions, containing a single modality. Using a pre-trained VLM, extended with CLAP for audio, we obtain the query and target embeddings, \( (\mathbf{h}_{q^+}, \mathbf{h}_{t}) \), by extracting the final-layer vector representation of the last token and train the embedding model with the standard InfoNCE loss~\cite{DBLP:journals/corr/abs-1807-03748}, incorporating in-batch negatives and hard negatives.
To enhance its multimodal retrieval capability, the training of the retrieval module is divided into two stages:

\begin{enumerate}
    \item \textbf{Text \& Image-Based Retrieval} \\
    In the first stage, the model is trained without direct audio input. Instead, audio is replaced by its corresponding captions, and only the session’s chat history and images are included for tuning. This allows the model to focus on retrieving memory based on textual and visual context.
    \item \textbf{Audio-Enhanced Retrieval} \\
    In the second stage, similar to the Dialogue Module, we integrate the CLAP with a linear adapter to enable direct audio comprehension. By incorporating audio embeddings, the retrieval module learns to process and retrieve memory using full multi-modal context, including raw audio information.
\end{enumerate}
This two-stage training approach ensures that the retrieval module effectively aligns and retrieves multi-modal representations, improving memory recall and interaction coherence across sessions. For detailed training settings, please refer to the section below.

\begin{figure}[t]
    \centering
      \includegraphics[width=0.90\linewidth]{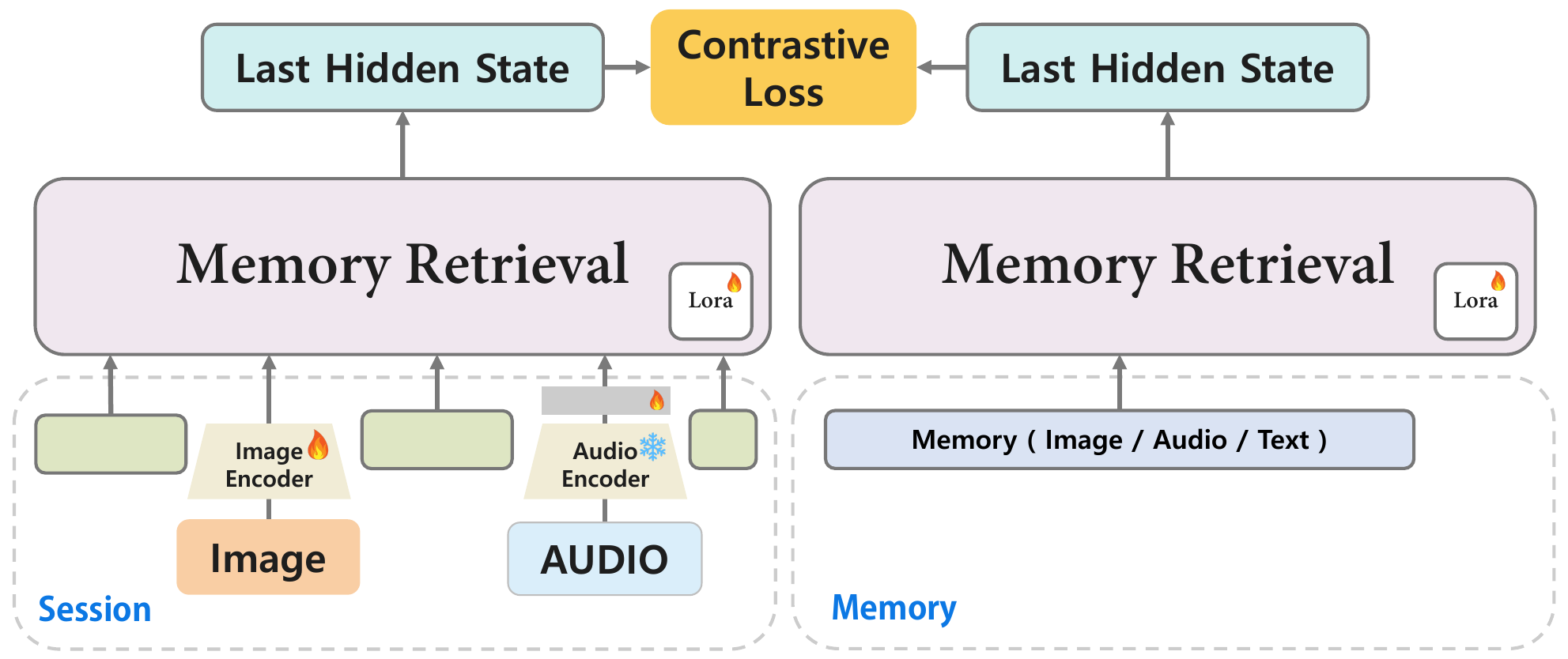}
    \caption{Overview of the retrieval module architecture, which utilizes contrastive loss for memory retrieval.}
    \label{fig:ret}
\end{figure}
\paragraph{Training Setting.}
We apply the LoRA to fine-tune Qwen2-VL-2B-Instruct, training with a cross-entropy loss, a maximum input length of 1024, and a learning rate of $1\times10^{-3}$. We use early stopping with a maximum of 3 epochs. For LoRA configuration, we set $r=8$, $\alpha=16$, dropout=0.05, and apply it to the q\_proj and v\_proj layers. During training stage 1, we use a batch size of 32, while in stage 2, where the audio adapter is trained, we reduce the batch size to 16. Training is conducted on 8 NVIDIA RTX A6000 GPUs. 

\section{Human Evaluation Details}
\label{sec:metrics}
We provide evaluators with guidelines, as shown in Table~\ref{tab:HuEvalData}, which include descriptions of the evaluation metrics and scoring criteria. Evaluators study these guidelines and complete a pilot test administered by the evaluation agency to ensure they fully understand the criteria before participating in the evaluation process. Additionally, Table~\ref{tab:FullEval} presents detailed human evaluation scores, including the average scores by group and inter agreement scores.

\section{Model Examples}
\label{sec:modelexam}
Please refer to Table~\ref{tab:MoToMo} for an example of conversation among models.

\section{Example of Model Comparison}
\label{sec:ExamMode}
Our model enables all speakers to engage in immersive, simultaneous conversations within a shared spatial and temporal context, based on the same modality input. Table~\ref{tab:imgconv2} and Table~\ref{tab:audioconv} show how our model's responses differ from those of existing models when processing visual and audio inputs, respectively.

\begin{table}[h]
\centering
\resizebox{0.95\linewidth}{!}{
\begin{tabular}{m{0.99\linewidth}}
\toprule
Generate a caption describing the image.\\
The caption should begin directly with details about the main objects and the situation.\\
Avoid starting with phrases like ``In this image''.\\
Write the caption from the observer's perspective, focusing solely on the given scene without considering other possible contexts.\\
\bottomrule
\end{tabular}
}
\caption{Prompt for caption revision.}
\label{tab:PromCaption}
\end{table}

\begin{table}[t]
\centering
\resizebox{0.95\linewidth}{!}{
\begin{tabular}{m{0.99\linewidth}}
\toprule
\textbf{For image}\\
\midrule
\#\#\#Instruction:\\
1. Given an image caption, identify the location using a single, general word.\\
2. Ensure the location is broadly applicable and captures the essence of the description.\\
3. If the specific location is unclear, make an educated guess based on where the described elements are typically found.\\
4. For example, if objects like a refrigerator, stove, and cooking utensils are visible, the location would be ``kitchen''. Similarly, a desk, computer, and books might suggest ``office''.\\
5. Avoid adding any additional explanations, introductions, or conclusions.\\
\\
\#\#\#Caption:\\
\{CAPTION\}\\
\\
\#\#\#Answer:\\
\\
\midrule
\textbf{For Audio}\\
\midrule
\#\#\#Instruction:\\
1. Given an audio caption, identify the location using a single, general word.\\
2. Ensure the location is broadly applicable and captures the essence of the description.\\
3. If the specific location is unclear, output ``none''.\\
4. If the sounds are fictional, unrealistic, or not commonly associated with real-world locations, output ``none''.\\
5. For example, sounds like sizzling, a refrigerator hum, and chopping might suggest ``kitchen''.\\
6. Avoid adding any additional explanations, introductions, or conclusions.\\
\\
\#\#\#Caption:\\
\{CAPTION\}\\
\\
\#\#\#Answer:\\
\\
\bottomrule
\end{tabular}
}
\caption{Prompt for specifying location based on caption.}
\label{tab:PromLocation}
\end{table}

\begin{table*}[t]
\centering
\resizebox{0.95\linewidth}{!}{
\begin{tabular}{m{0.99\linewidth}}
\toprule
\#\#\#Instruction:\\
1. Four speakers are involved in a conversation episode, consisting of one main speaker and three partners. Each session will feature the main speaker alongside two different partners.\\
2. There will be three continuous sessions. In each session, the main speaker and two different partners will witness two settings unfolding in real-time right in front of them. These settings are live observations, experienced not only visually but also audibly, and are not based on past memories, images, or pictures.\\
3. From a provided list of settings, select two unique settings for each session. Ensure that no setting is repeated across the sessions.\\
4. There will be a time gap between each session. Choose suitable time intervals from the following options: ``a few hours later,'' ``a few days later'', ``a few weeks later'', ``a few months later'', and ``a couple of years later''.\\
5. Clearly define the names and relationships of all speakers to provide context for their interactions and to enhance the flow of conversation.\\
6. During each session, the speakers should engage in discussions that logically connect to the context of the settings they observe, without directly referencing specific details of those settings. For instance, if the setting involves cooking, one might inquire about the dish being prepared.\\
7. Ensure that the combination of settings in each session does not include contradictory elements, such as differing weather conditions or inconsistent times of day that would not logically coexist.\\
8. Please do not generate any other opening, closing, and explanations.\\
\\
\#\#\#Setting list:\\
- \{MODALITY LIST\}\\
\\
\#\#\#Response:\\
- Main speaker name: \{insert name\}\\
- Main speaker relationship: \{insert relationship\}\\
- Partner 1 name: \{insert name\}\\
- Partner 1 relationship: \{insert relationship\}\\
- Partner 2 name: \{insert name\}\\
- Partner 2 relationship: \{insert relationship\}\\
- Partner 3 name: \{insert name\}\\
- Partner 3 relationship: \{insert relationship\}\\
- Scene numbers for session 1: \{insert scene number 1\}, \{insert scene number 2\}\\
- Two partners' names in Scene 1: \{insert partner name 1\}, \{insert partner name 1\}\\
- Time interval between session 1 and 2: \{insert time interval\}\\
- Scene numbers for session 2: \{insert scene number 1\}, \{insert scene number 2\}\\
- Two partners' names in Scene 2: \{insert partner name 1\}, \{insert partner name 1\}\\
- Time interval between session 2 and 3: \{insert time interval\}\\
- Scene numbers for session 3: \{insert scene number 1\}, \{insert scene number 2\}\\
- Two partners' names in Scene 3: \{insert partner name 1\}, \{insert partner name 1\}\
\\
\bottomrule
\end{tabular}
}
\caption{Prompt for scenario generation.}
\label{tab:PromCandidate}
\end{table*}

\begin{table}[t]
\centering
\resizebox{0.95\linewidth}{!}{
\begin{tabular}{m{0.99\linewidth}}
\toprule
\#\#\#Instruction:\\
1. Two captions are provided, each describing either an image or an audio clip.\\
2. Determine whether the two captions are aligned and compatible.\\
3. Assume that the speakers are observing or listening to the described image or audio in real-time within the same context.\\
4. If the two captions cannot coexist within the same context, output ``no'' Otherwise, output ``yes''.\\
5. For example, if one caption describes a ski resort and the other describes a meadow, the two captions cannot coexist, and the output should be ``no''.\\
6. Please do not generate any other opening, closing, and explanations.\\
\\
\#\#\#Caption:\\
- \{CAPTION A\}\\
- \{CAPTION B\}\\
\\
\#\#\#Answer:\\
\\
\bottomrule
\end{tabular}
}
\caption{Prompt for checking modality alignment within a session.}
\label{tab:PromAlign}
\end{table}

\begin{table}[t]
\centering
\resizebox{0.95\linewidth}{!}{
\begin{tabular}{m{0.99\linewidth}}
\toprule
\#\#\#First session conversation:\\
\{FIRST SESSION\}\\
\\
\#\#\#Second session conversation:\\
\{SECOND SESSION\}\\
\\
\#\#\#Instruction:\\
The following is a third-session conversation, \{TIME INTERVAL\} the second-session conversation, between \{MAIN SPEAKER NAME\} (\{MAIN SPEAKER RELATIONSHIP\}), \{PARTNER 1 NAME\} (\{PARTNER 1 RELATIONSHIP\}) and \{PARTNER 2 NAME\} (\{PARTNER 2 RELATIONSHIP\}).\\
\\
During the conversation, \{MAIN SPEAKER NAME\}, \{PARTNER 1 NAME\} and \{PARTNER 2 NAME\} participate in the following settings in real-time:\\
- \{CAPTION 1\}\\
- \{CAPTION 2\}\\
\\
The following speakers, \{MAIN SPEAKER NAME\}, \{PARTNER 1 NAME\} and \{PARTNER 2 NAME\}, participate in the settings in real-time during the conversation, so we avoid directly describing the settings. Instead, we proceed with the dialogue in a way that shows they are sharing the same settings. We also avoid discussing impressions, feelings, and imagination.\\
For example, if there is a cutting board and cooking ingredients in the kitchen, we could continue the conversation by asking what kind of dish to make. We maintain the natural flow of the dialogue, not just talking about settings.\\
Especially, please write the dialogue with the understanding that it is a continuation of the previous sessions. Feel free to reference the previous sessions as needed, and consider the time interval between the sessions. Avoid forcing connections if they don't naturally flow; the dialogue should reflect the continuity without feeling contrived.\\
The conversation should begin with \{PARTNER 1 NAME\} or \{PARTNER 1 NAME\} initiating it. Also, each speaker's statements start with their name in brackets. For example, \{PARTNER 1 NAME\}'s statements start with ``[\{PARTNER 1 NAME\}]''.\\
The order of the speakers' utterance is determined randomly. There is no need to follow a fixed order when speaking.\\
Complete the conversation in exactly that format.\\
\bottomrule
\end{tabular}
}
\caption{Prompt for session conversation generation.}
\label{tab:PromSession}
\end{table}

\begin{table}[t]
\centering
\resizebox{0.95\linewidth}{!}{
\begin{tabular}{m{0.99\linewidth}}
\toprule
\#\#\#Instruction:\\
1. Two settings and a list of dialogue utterances are provided.\\
2. Based on the dialogue, identify the utterance number where each of the two settings is first mentioned, seen, or heard by the speakers in real-time, including both visual and auditory elements.\\
3. Select only one utterance number for each setting.\\
4. Please do not generate any other opening, closing, and explanations.\\
\\
\#\#\#Settings:\\
- \{CAPTION A\}\\
- \{CAPTION B\}\\
\\
\#\#\#Utterances:\\
\{UTTERANCE LIST\}\\
\\
\#\#\#Response:
\\
\bottomrule
\end{tabular}
}
\caption{Prompt for modality tagging.}
\label{tab:PromTagging}
\end{table}

\begin{table*}[t]
\centering
\resizebox{0.95\linewidth}{!}{
\begin{tabular}{m{0.99\linewidth}}
\toprule
\#\#\#Second session conversation:\\
- First Setting: \{FIRST SETTING\}\\
- Second Setting: \{SECOND SETTING\}\\
\{SESSION CONVERSATION\}\\
\\
\#\#\#Instruction:\\
1. Please summarize the conversation from \{MAIN SPEAKER NAME\}'s perspective. The summary should focus on what \{MAIN SPEAKER NAME\} needs to remember for the next conversation.\\
2. The summary should include emotions, thoughts, facts, and commitments expressed during the conversation, but only those explicitly revealed during the discussion. Exclude general descriptions or background information about the setting.\\
3. Do not include content from the setting descriptions in the memory. Only include information shared during the conversation.\\
4. Ensure the memory entries are concise and focus only on unique, relevant information necessary for the next session. Each memory entry must focus on one person only, without combining multiple speakers into one sentence or perspective.\\
5. Each memory entry should be a separate sentence or key idea, but avoid summarizing every single statement unless it is crucial for the next session. All memory entries must be separated by a ``/''. For example, ``\{\{sentence 1\}\} / \{\{sentence 2\}\} ...''.\\
6. The memory must be presented from \{MAIN SPEAKER NAME\}'s perspective, focusing on summarizing the overall key points or themes rather than individual conversational details.\\
7. Avoid duplicating information already included in the summary. Consolidate similar points into a single entry if possible.\\
8. If a memory element updates or replaces existing information, ensure the updated version reflects the latest understanding and avoid redundancy.\\
9. When summarizing \{MAIN SPEAKER NAME\}'s memory about themselves, start the memory entry with ''I ~''.\\
10. At the end of each memory sentence, specify who the memory is about from \{MAIN SPEAKER NAME\}'s perspective using the format ``(about \{\{name\}\})\\. For example, ``\{\{Memory Sentence\}\} (about \{\{name\}\})''. Ensure that the parentheses end with a period to complete the sentence.\\
11. Do not use group references like ``(about all)'' or combine multiple people into one memory entry.\\
12. If there are no memories to summarize, output ``no memory''.\\
13. Avoid adding extra explanations, introductions, or conclusions.\\
\\
\#\#\#Response:
\\
\bottomrule
\end{tabular}
}
\caption{Prompt for memory generation.}
\label{tab:PromMemGen}
\end{table*}

\begin{table}[t]
\centering
\resizebox{0.95\linewidth}{!}{
\begin{tabular}{m{0.99\linewidth}}
\toprule
\#\#\#First session setting:\\
1 - \{CAPTION\}\\
2 - \{CAPTION\}\\
\\
\#\#\#Second session setting:\\
3 - \{CAPTION\}\\
4 - \{CAPTION\}\\
\\
\#\#\#\{MAIN SPEAKER NAME\}'s memory from first session:\\
\{FIRST SESSION MEMORY\}\\
\\
\#\#\#\{MAIN SPEAKER NAME\}'s memory from second session:\\
\{SECOND SESSION MEMORY\}\\
\\
\#\#\#Instruction:\\
1. Summarized conversation memories from the perspective of \{MAIN SPEAKER NAME\} and the settings used in the dialogue are provided.\\
2. \{MAIN SPEAKER NAME\} references the settings and memories from previous sessions during subsequent conversations to ensure seamless continuity in the dialogue.\\
3. When referring to memories, the goal is to provide rich context by connecting related elements.\\
4. If the provided settings and memories pertain to the same context or reflect subsequent updates, they should be connected.  \\
5. The format for expressing connections is ``\{\{NUMBER\}\}-\{\{NUMBER\}\}''and should be output on separate lines.\\
6. Each \{\{NUMBER\}\} should contain only a single digit. When multiple memories are connected, write each connection on a new line.\\
7. Avoid adding extra explanations, introductions, or conclusions.\\
\\
\#\#\#Response:
\\
\bottomrule
\end{tabular}
}
\caption{Prompt for memory linking.}
\label{tab:PromMemLink}
\end{table}

\begin{table}[t]
\centering
\resizebox{0.95\linewidth}{!}{
\begin{tabular}{m{0.99\linewidth}}
\toprule
\#\#\#Third session conversation:\\
\{SESSION CONVERSATION\}\\
\\
\#\#\#\{MAIN SPEAKER NAME\}'s memory:\\
\{MEMORY LIST\}\\
\\
\#\#\#Instruction:\\
1. \{MAIN SPEAKER NAME\} engages in a third session conversation based on their memory.\\
2. The utterances by \{MAIN SPEAKER NAME\} are informed by their referenced memory.\\
3. During the conversation, if any part of \{MAIN SPEAKER NAME\}'s utterances relies on memory, that part must be specifically identified.\\
4. The memory consists of two components: (1) Summarized information from previous session conversations, and (2)Real-time observations by \{MAIN SPEAKER NAME\} during prior sessions.\\
5. The output format should follow the structure: ``Utterance Letter-Memory Number'' (e.g., ``A-3'').\\
6. Each line should contain a single ``Utterance Letter-Memory Number'' entry, separated by a newline.\\
7. If an utterance references multiple memory entries, each reference must be listed on a new line rather than combining memory numbers on the same line.\\
8. If the same information needs to be referenced, prioritize using the memory from the most recent session.\\
9. If no memory is referenced and there is nothing to output, return ``none''.\\
10. Avoid adding extra explanations, introductions, or conclusions.\\
\\
\#\#\#Response:
\\
\bottomrule
\end{tabular}
}
\caption{Prompt for memory tagging.}
\label{tab:PromMemTag}
\end{table}

\begin{table*}[t]
\centering
\resizebox{0.95\linewidth}{!}{
\begin{tabular}{m{0.99\linewidth}}
\toprule
\#\#\#Conversation episode:\\
\\
* First session:\\
\{FIRST SESSION\}\\
\\
* Second session:\\
\{SECOND SESSION\}\\
\\
* Third session:\\
\{THIRD SESSION\}\\
\\
\#\#\#Instruction:\\
- Each conversation episode consists of three sessions, with one main speaker and three conversation partners.\\
- In each session, the main speaker converses with two conversation partners, and the partner combinations may vary between sessions.\\
- During each session, the speakers either observe or participate in two real-time settings.\\
- The settings being observed are not images, photos, or past memories but are happening live in front of them, with both visual and auditory elements unfolding in real time.\\
- As a judge, you will evaluate the conversation episode based on the following criteria.\\
- Please refrain from adding any other introductions, conclusions, or explanations.\\
\\
\#\#\#Metric:\\
1. Is there complete consistency between the environmental, spatial, and temporal features of the settings within the session? For example, it would be contradictory if one setting depicts daytime while the other depicts nighttime, or if spatial features (e.g., location or layout) and time progression are logically inconsistent. (Yes or No)\\
2. Do all sessions within the episode maintain a plausible continuity in time, space, and context? For example, any stated time intervals or implied transitions between settings should be logical and coherent. (Yes or No)\\
3. Are all participants depicted as fully engaging with the setting in real time? References to past or future events should not imply detachment from the present interaction (e.g., avoiding phrases like ``for our next scene'' or references to reviewing recorded footage). (Yes or No)\\
4. Are all settings within the session entirely realistic? Any elements that seem exaggerated, cartoonish, or overly stylized for natural conversation or interaction should be avoided. (Yes or No)\\
5. Is each setting fully utilized and referenced in the conversation? All settings presented within the session must have a clear role in the dialogue or interaction, without any being neglected. (Yes or No)\\
6. Do all spoken lines reflect the tone and context of natural, real-time interaction? For instance, lines should avoid referring to the setting or events in a way that suggests they are pre-recorded, staged, or viewed from an external perspective. (Yes or No)\\
\\
\#\#\#Response:\\
1.\\
2.\\
3.\\
4.\\
5.\\
6.\\
\bottomrule
\end{tabular}
}
\caption{Prompt for episode validation.}
\label{tab:PromDataValid}
\end{table*}

\begin{table*}[t]
\centering
\resizebox{0.99\linewidth}{!}{
\begin{tabular}{l p{15cm} }
\toprule
\multicolumn{2}{l}{\bf Session 1}\\
\midrule
& \includegraphics[height=3.5cm]{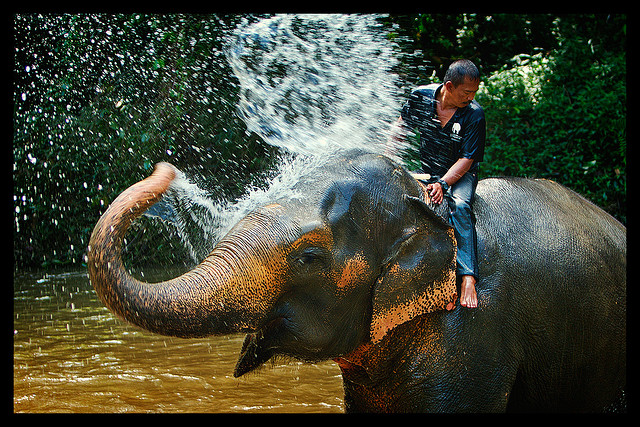} \\
\textbf{Jamie} & Did you see how the elephant just sprayed water? It's like it's having a blast!\\
\textbf{Taylor} & Right? And look at the rider; he’s handling that balance like a pro. I wouldn’t want to fall into that river.\\
\textbf{Alex} & I can only imagine how muddy it must be out there! Bet the rider picked the perfect time to hop on.\\
\textbf{Jamie} & I wonder if they had to train for this. It looks like a lot of fun, but maintaining that balance can’t be easy.\\
& \includegraphics[height=3.5cm]{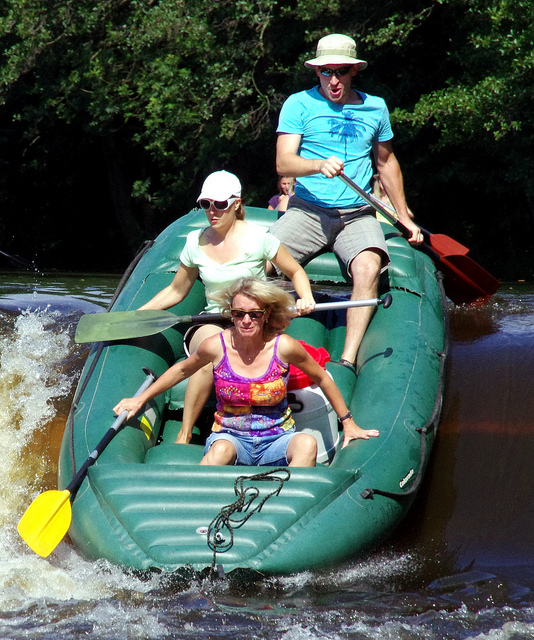} \\
\textbf{Taylor} & Speaking of fun, check out those guys on the raft! They’re about to hit that small waterfall.\\
\textbf{Alex} & They look a bit tense, but it seems like they’re ready for the rush. I hope they paddle right!\\
\textbf{Jamie} & I can’t tell if they’re nervous or excited. That’s such a perfect day for a rafting adventure, though.\\
\textbf{Taylor} & Definitely! It feels like a great day for both activities. I’d love to try that raft ride sometime.\\
\textbf{Alex} & Me too! But only if I can guarantee not falling out. That water looks pretty wild.\\
\textbf{Jamie} & And you know how slippery it can get! But I think with a good crew, we could make it work.\\
\textbf{Taylor} & Yeah, teamwork makes the dream work. Just like that elephant and rider out there.\\
\textbf{Alex} & Now that’s a solid comparison! Might be more at risk of muddy water than a tidal wave, though.\\
\textbf{Jamie} & True! Plus, no one would want a trunkful of water splashed on them while they're navigating!\\
\textbf{Taylor} & Right! But at the end of the day, it all just adds to the fun. Both experiences seem pretty unforgettable.\\
\bottomrule
\end{tabular}
}
\caption{An example of the first session in the dataset episode.}
\label{tab:DataExam1}  
\end{table*}

\begin{table*}[t]
\centering
\resizebox{0.99\linewidth}{!}{
\begin{tabular}{l p{15cm} }
\toprule
\multicolumn{2}{l}{\bf Session 2}\\
\midrule
& \includegraphics[height=3.5cm]{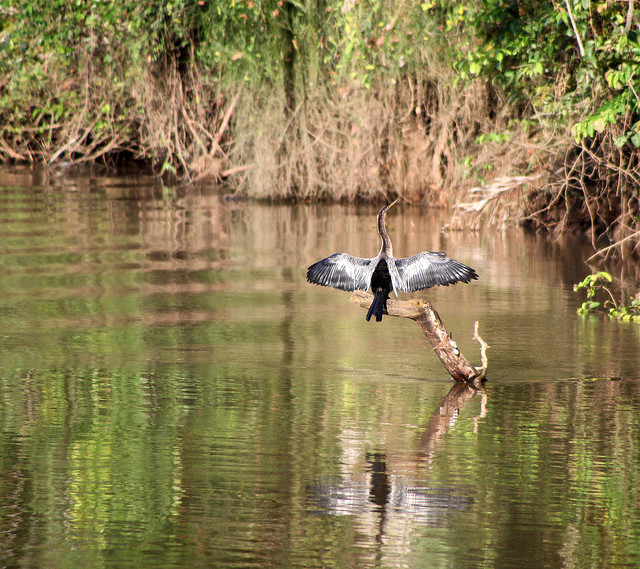} \\
\textbf{Jamie} & That cormorant over there looks like it's ready for a photoshoot! Its reflection is almost perfect on the water.\\
\grayrow \multicolumn{2}{c}{\bf Clotho dataset - 39935.wav}\\
\textbf{Taylor} & I can’t believe how still the water is right now. It's like a painting. But just wait until that motorboat speeds by!\\
\textbf{Alex} & Right? The contrast between the calmness of the river and the chaos of the engine is pretty interesting.\\
\textbf{Jamie} & Totally! Kind of like how we were all raving about the rafting and the elephant the other day. Different vibes, but equally fun!\\
\textbf{Taylor} & Speaking of fun, I wonder if those folks on the boat are going to go full throttle as soon as they hit open water.\\
\textbf{Alex} & They probably will! I mean, who wouldn’t want to crank up the excitement with some splashes like that?\\
\textbf{Jamie} & It's like they’re showcasing their own version of adventure. You know, I’d still prefer the elephant ride for a more chilled experience.\\
\textbf{Taylor} & Fair point! But the adrenaline rush from that motorboat sounds tempting too. I’d want to try both!\\
\textbf{Alex} & Agreed. We should plan a day and do both! Maybe we can start with the raft ride, ease into the motorboat energy, and then watch those cormorants dry off.\\
\textbf{Jamie} & That sounds like a solid plan! Just hope the motorboat doesn’t stir up too much mud and ruin our view of the cormorant.\\
\textbf{Taylor} & True! But if it does, it’ll give us some splashes to laugh about. Plus, we’ll still have that serene moment before it arrives.\\
\textbf{Alex} & That balance of calm and excitement is what makes these adventures memorable, right?\\
\textbf{Jamie} & Exactly! Just like our last outing. It’s all about enjoying whatever comes our way!\\
\textbf{Taylor} & And capturing those moments! A bit of mud won’t matter when we’re having a blast.\\
\textbf{Alex} & Alright then, let’s keep an eye on the boat and the cormorant. It’s the best of both worlds!\\
\bottomrule
\end{tabular}
}
\caption{An example of the second session in the dataset episode.}
\label{tab:DataExam2}  
\end{table*}

\begin{table*}[t]
\centering
\resizebox{0.99\linewidth}{!}{
\begin{tabular}{l p{15cm} }
\toprule
\multicolumn{2}{l}{\bf Session 3}\\
\midrule
& \includegraphics[height=3.5cm]{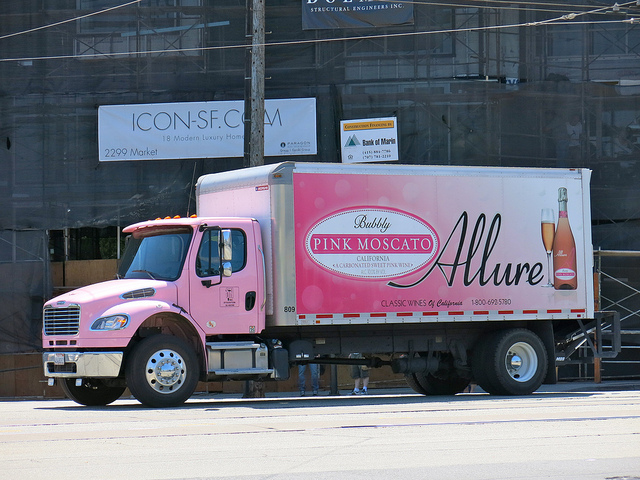} \\
\textbf{Taylor} & Look at this truck! I can’t believe they’ve got a whole advertisement for Moscato right here. It’s like they’re throwing a party in the middle of the street!\\
& \includegraphics[height=3.5cm]{Image/000000145480.jpg} \\
\textbf{Alex} & Right? It really stands out with that bright pink color. Not to mention that candle on the parking meter—it’s like an indoor celebration outdoors!\\
\textbf{Morgan} & Totally! It’s a fun touch. Speaking of parties, does this give anyone else an idea for a get-together? Maybe we could have a little wine tasting!\\
\textbf{Alex} & A wine tasting sounds perfect! We could grab a few different types. I remember our last adventure—we really enjoyed trying new things together.\\
\textbf{Taylor} & Yes! And we can recreate the vibe with some snacks. Maybe some cheese and crackers to go along with the wine.\\
\textbf{Morgan} & Great idea! I’d also suggest we include some fruit pairs—grapes and strawberries could complement the Moscato well.\\
\textbf{Alex} & Definitely! By the way, it’s been a while since our last outing. Remember that rafting trip? I think we should make this tasting our next adventure!\\
\textbf{Taylor} & Awesome plan! First, the wine tasting, then we can decide on our next outdoor adventure. Who knows, maybe we'll find some activity that even tops the elephant ride!\\
\textbf{Morgan} & Sounds exciting! Just think of the great stories we would have. Plus, with the candle on the meter, it feels like we already have a fun theme going.\\
\textbf{Alex} & Right? A whimsical birthday party vibe! We could even get some decorations to add to it.\\
\textbf{Taylor} & We’ll have to check if we can get some balloons too. It would be perfect! You know, make it festive.\\
\textbf{Morgan} & Yes! And who knows, maybe we’ll attract a few cheerful people passing by. After all, who doesn’t love a celebration?\\
\textbf{Alex} & Exactly! I say we take a few pictures to document the moment, just like we did when we spotted that cormorant the other day.\\
\textbf{Taylor} & Great call! Those memories are always fun to look back on. Plus, it’ll give us a reason to gather again soon!\\
\textbf{Morgan} & It’s settled then! Wine tasting with a little playful theme and photo moments sounds like a fantastic idea!\\
\textbf{Alex} & I’m in! Let’s plan it out. I can already envision some memorable stories coming out of it, just like our previous outings.\\
\textbf{Taylor} & Can’t wait! Let’s do it!\\
\bottomrule
\end{tabular}
}
\caption{An example of the third session in the dataset episode.}
\label{tab:DataExam3}  
\end{table*}

\begin{table*}[t]
\centering
\resizebox{0.99\linewidth}{!}{
\begin{tabular}{l p{15cm} }
\toprule
\multicolumn{2}{l}{\bf Example Session}\\
\midrule
& \includegraphics[height=3.5cm]{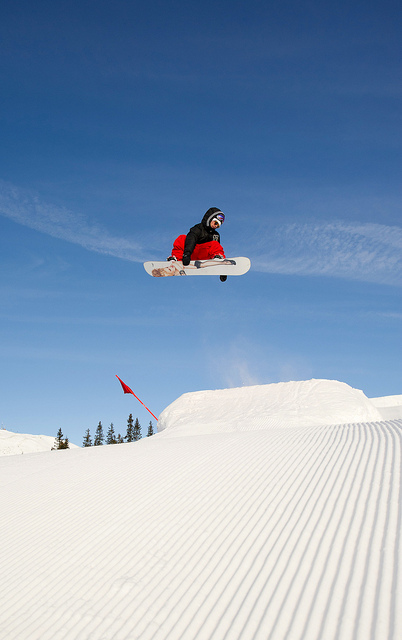} \\
\textbf{Jamie} & Did you see that snowboarder? They really nailed that jump!\\
\textbf{Sam} & Yeah, it looked awesome! The way they soared above the slope was incredible. I wish I could do that.\\
\textbf{Alex} & It takes a lot of practice to get those air tricks down. I just love watching the way they spin mid-air.\\
\textbf{Jamie} & For sure! And it seems like the snow conditions are perfect today, too. Really nice and packed.\\
\textbf{Sam} & I wonder how the skiers are faring. That jump off the slope was impressive as well.\\
\textbf{Alex} & Definitely! I noticed that skier launched off with so much speed. The sun must be ideal for those fast runs.\\
\textbf{Jamie} & I hope they have good control on that steep descent. It’s always thrilling to watch them tackle those angles.\\
& \includegraphics[height=3.5cm]{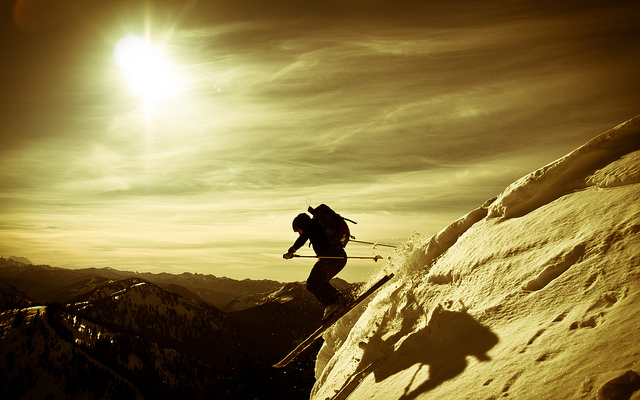} \\
\textbf{Sam} & It’s interesting how different the techniques are between skiing and snowboarding. One is all about edges, while the other is about balance.\\
\textbf{Alex} & Right! Each sport has its own style. I think that blending both could create a unique trick.\\
\textbf{Jamie} & Speaking of tricks, we should try to record some of these stunts. It would be great to have a video to look back on later!\\
\textbf{Sam} & Great idea! Even if we can't perform them ourselves, at least we can capture the thrill of it all.\\
\textbf{Alex} & Let’s pick a spot that has a good view of both the skiers and snowboarders.\\
\textbf{Jamie} & Perfect! I’ll grab my phone. It’ll be fun to have some footage to remember this day.\\
\textbf{Sam} & I'm in! Maybe we can even get a few shots of those stylish wipeouts—those are always entertaining.\\
\textbf{Alex} & Absolutely, let’s not forget to cheer them on, too. It makes it more exciting for everyone while we film!\\
\midrule
\multicolumn{2}{l}{\bf Summarized Memory}\\
\midrule
1. & I enjoy watching snowboarders and skiers perform tricks in the snow. (from first session, about me)\\
2. & Jamie suggested recording some stunts, and I think it would be fun to have footage to remember this day. (from first session, about Jamie)\\
3. & Sam is interested in capturing wipeouts, which adds an entertaining element to our video. (from first session, about Sam)\\
4. & I believe that each sport has its own unique style and that blending both could create something interesting. (from first session, about me)\\
5. & It's important to cheer on the performers while filming, as it creates a more exciting atmosphere. (from first session, about me)\\
\bottomrule
\end{tabular}
}
\caption{An example of a memory summarized from the perspective of the main speaker (Alex) in a conversation. The summarized memory is combined with modality elements from the session to form a multimodal memory.}
\label{tab:DataMemSum}  
\end{table*}

\begin{table}[t]
\centering
\resizebox{0.95\linewidth}{!}{
\begin{tabular}{m{0.99\linewidth}}
\toprule
\grayrow \textbf{Emergency vehicle siren goes off for the entire time}\\
\grayrow \textbf{Clotho dataset - 44772.wav}\\
I noted that there was a lot going on that night, including the rally and the lights along the pathway. (from first session, about me)\\
\midrule
\includegraphics[height=3.5cm]{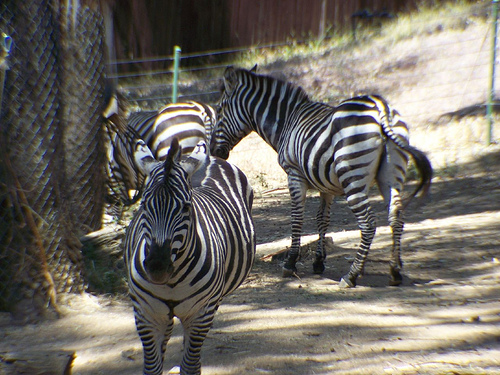}\\
Observing the zebras revealed to me that different species have their own unique ways of relaxing in their environments (about Jamie)\\
\midrule
\includegraphics[height=3.5cm]{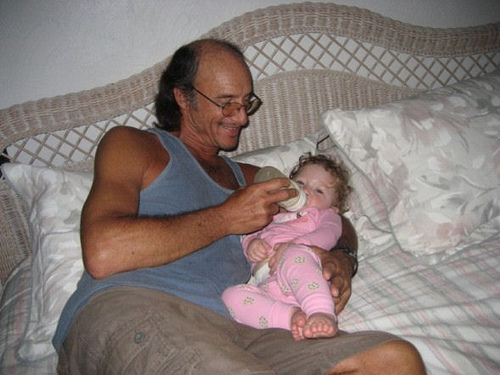}\\
\includegraphics[height=3.5cm]{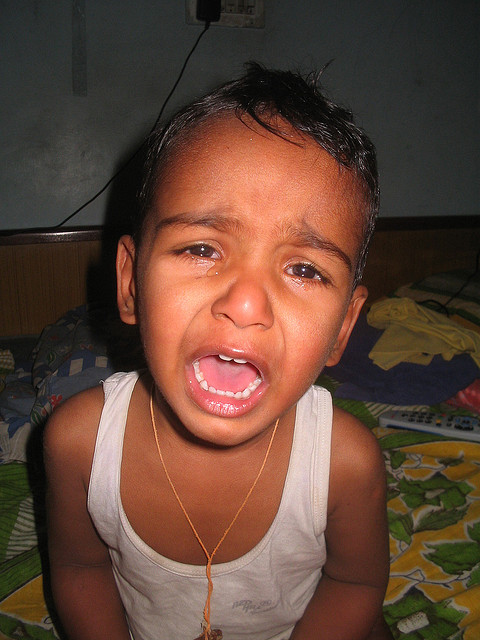}\\
\bottomrule
\end{tabular}
}
\caption{Examples of multimodal memory links. The first row shows links between audio clips and text memories, the second row shows links between images and text memories, and the third row shows links between two different modalities.}
\label{tab:DataMemLink}
\end{table}
\begin{table*}[t]
\centering
\resizebox{0.95\linewidth}{!}{
\begin{tabular}{m{0.99\linewidth}}
\toprule
\textbf{Multimodal Conversation Evaluation}\\
\\
You are asked to evaluate a multimodal conversation, which integrates multiple modalities such as images or audio within the dialogue. Each conversation episode consists of three consecutive sessions, with two modalities featured in each session. Each episode involves four speakers in total, with three randomly selected to participate in any given session. Your evaluation should focus on both the overall quality of the conversation and the effectiveness with which the modalities are reflected in the dialogue.\\
\\
\grayrow * For Dataset:\\
\textbf{(1) coherence and consistency}\\
Is the conversation coherent throughout, with no contradictions or logical inconsistencies?\\
\\
\textbf{(2) modalities alignment}\\
Do the modalities in a single session align within the same context of place and time? For example, if one modality features an image of a winter ski resort and another modality includes the sound of a beach ocean, that would be considered inconsistent and incorrect.\\
\\
\textbf{(3) modalities engagement}
Are the presented modalities (image, audio) appropriately reflected in the dialogue, and do the speakers actively engage with the modalities in real time?\\
\\
\textbf{(4) memorability}\\
Do conversations in subsequent sessions effectively recall and build upon the content of previous sessions, considering the time intervals between them? Additionally, does the memory summary of the first and second sessions accurately reflect the content of those conversation?\\
\\
\grayrow * For Model:\\
\textbf{(1) naturalness}\\
Do all speakers demonstrate fluency and natural conversational flow, with smooth transitions between dialogue (turns, sessions) and seamless incorporation of modality cues (images and audio) without awkward pauses or disruptions?\\
\\
\textbf{(2) Immersive}\\
Do the speakers naturally engage with the presented modalities (images and audio) as if they are experiencing them simultaneously in the same spatial and temporal environment, integrating them seamlessly into the conversation to create a realistic and contextually rich dialogue?\\
\\
\textbf{(3) memorability}\\
Do the speakers accurately recall and incorporate relevant information from previous sessions, including both textual and multimodal (visual and auditory) memory, using this information appropriately to maintain context and enhance the flow of the conversation?\\
\\
\grayrow * Possible Options:\\
Bad / Slightly Bad / Neutral / Slightly Good / Good\\
\\
A top score of 5 is reserved for conversations that are entirely flawless, while a score of 1 is assigned when severe issues make a proper evaluation impossible. A rating of 4 indicates an overall excellent performance, whereas a score of 3 reflects the presence of only minor imperfections. A score of 2 is given when notable issues are detected, even though they do not completely undermine the conversation’s quality.\\
\bottomrule
\end{tabular}
}
\caption{Human evaluation guidelines containing descriptions of metrics and criteria.}
\label{tab:HuEvalData}
\end{table*}
\begin{table*}[t]
\centering
\resizebox{0.80\linewidth}{!}{
\begin{tabular}{c c c c c c c}
\hline
\textbf{Metric} & \textbf{Group 1} & \textbf{Group 2} & \textbf{Group 3} & \textbf{Average} & \textbf{Agreement} & \textbf{Machine}\\
\hline
\grayrow\multicolumn{7}{c}{\textbf{Dataset Quality}} \\
Coherence and Consistency & 4.84 & 4.85 & 4.73 & 4.81 & 0.54 & 4.99\\
Memorability & 4.56 & 4.79 & 4.54 & 4.63 & 0.30 & 4.99\\
Modalities Alignment & 4.13 & 4.33 & 4.18 & 4.21 & 0.27 & 4.26\\
Modalities Engagement & 4.37 & 4.48 & 4.23 & 4.36 & 0.32 & 4.57\\
\hline
\textbf{Dataset Overall} & 4.48 & 4.61 & 4.42 & 4.50 & - & 4.70\\
\hline
\grayrow\multicolumn{7}{c}{\textbf{Model Performance}} \\
Naturalness & 4.28 & 4.58 & 4.16 & 4.34 & 0.20 & 4.68\\
Immersive & 4.12 & 4.25 & 4.05 & 4.14 & 0.40 & 4.56\\
Memorability & 4.20 & 4.68 & 4.17 & 4.35 & 0.28 & 4.46\\
\hline
\textbf{Model Overall} & 4.20 & 4.50 & 4.13 & 4.28 & - & 4.57\\
\hline
\end{tabular}
}
\caption{Full evaluation results for both dataset quality and model performance, including group-specific scores and agreement. Agreement refers to the rate at which responses match across all conversations from all three groups.}
\label{tab:FullEval}
\end{table*}

\begin{table*}[t]
\centering
\resizebox{0.99\linewidth}{!}{
\begin{tabular}{l p{15cm} }
\toprule
\textbf{Alex} & Did you see that turn I just made? It felt perfectly sharp!\\
\textbf{Taylor} & I did! That’s impressive. That’s the key to leaning more, isn’t it?\\
\textbf{Morgan} & And it’s also about control, like you mentioned earlier. Those small adjustments can make a huge difference.\\
\textbf{Alex} & Absolutely! Just like with the way we prepare for a match, small adjustments can lead to big wins.\\
\textbf{Morgan} & It’s the same with skiing, isn’t it? A slight tweak here and there can make the difference between a perfect turn and a misstep.\\
\textbf{Alex} & Definitely! And what’s interesting is how we can learn from other skiers, just like we can from games or even sports.\\
& \includegraphics[height=3.5cm]{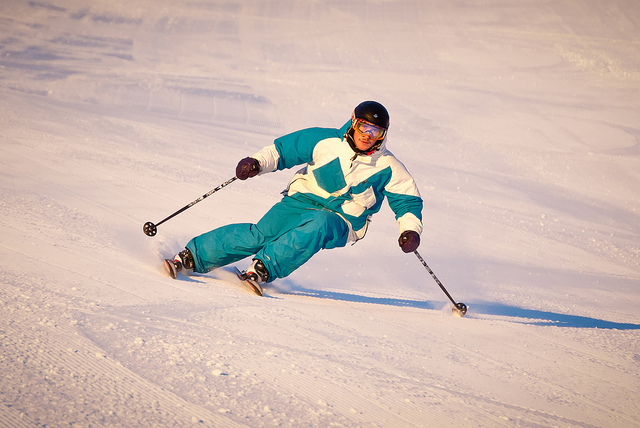} \\
\textbf{Morgan} & It seems they find the right line, don’t they? You really have to know how to read the snow.\\
\textbf{Taylor} & Right! And just as we need to adapt to the game, we might want to think of how to adapt to the skiing conditions.\\
\textbf{Alex} & Speaking of adaptation, I noticed that the skiers seem to change their speed and direction frequently. It’s fascinating how they navigate through those shifts.\\
\textbf{Morgan} & I wonder how they gauge their speed compared to the rest of the field, just like we must be attentive to our competition.\\
\bottomrule
\end{tabular}
}
\caption{Example of a conversation among three models.}
\label{tab:MoToMo}  
\end{table*}
\begin{table}[t]
\centering
\resizebox{0.99\linewidth}{!}{
\begin{tabular}{l p{7.5cm} }
\toprule
\textbf{User} & Wow, I'm so excited! Where should we go first?\\
& \includegraphics[height=3.5cm]{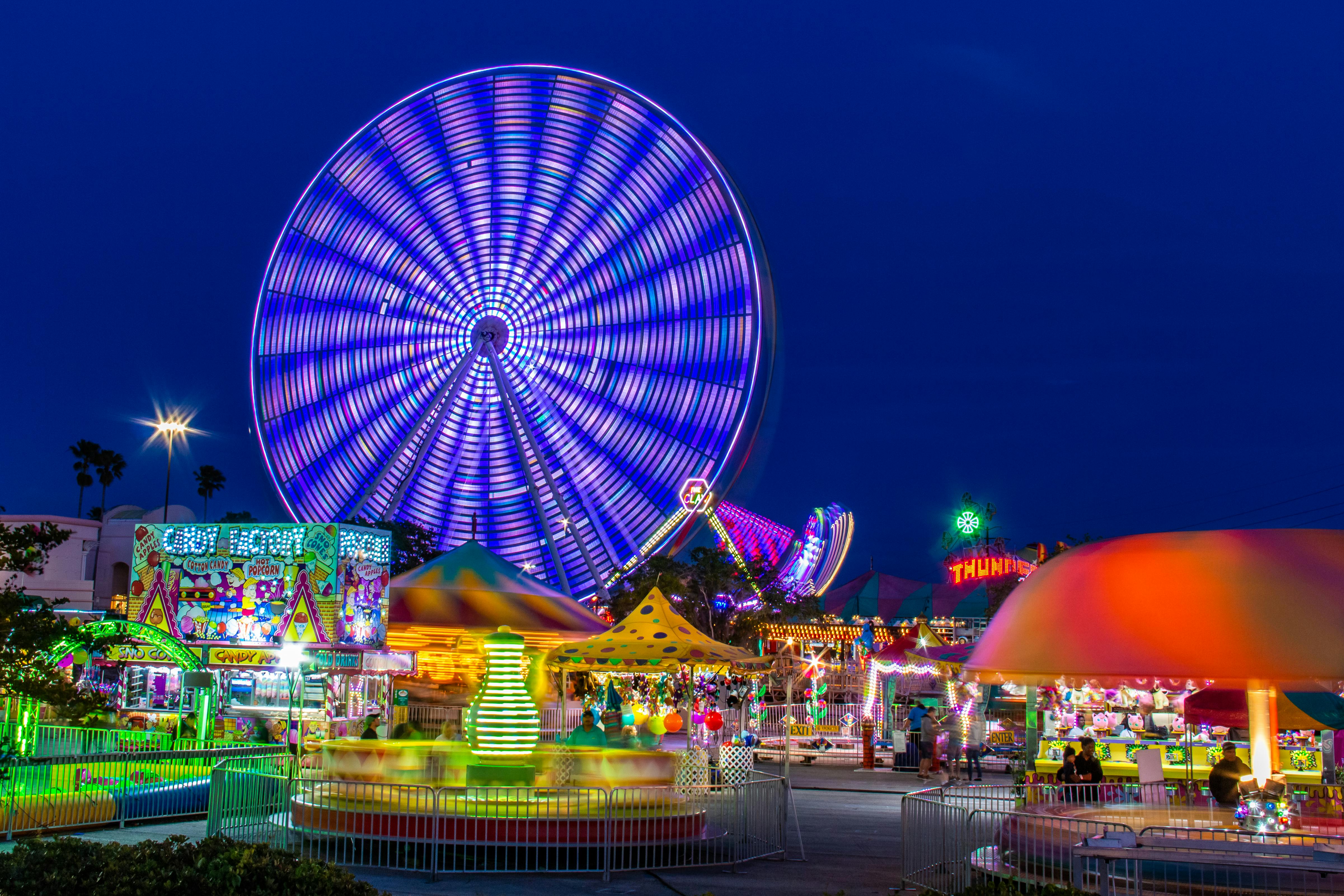} \\
\midrule
\textbf{Qwen2} & It looks like you're at an amusement park with lots of fun rides and attractions to choose from! Here’s how you might decide where to start: ...... \\
\midrule
\textbf{LLaVA} & The image shows a lively carnival with a large Ferris wheel and other rides, including a roller coaster. There are also several people walking around and enjoying the festive atmosphere ......\\
\midrule
\textbf{o3-mini} & Check out that giant Ferris wheel—it’s basically the park’s lighthouse! I’d kick things off there ......\\
\midrule
\textbf{Ours} & Is that ride near here? That could be a great idea! Let's check the map and see where it is.\\
\bottomrule
\end{tabular}
}
\caption{An example of a live human chat comparing how different models respond to image input.}
\label{tab:imgconv2}
\vspace{-7pt}
\end{table}

\begin{table}[t]
\centering
\resizebox{0.99\linewidth}{!}{
\begin{tabular}{l p{7.5cm} }
\toprule
\textbf{User} & Whoa! What just happened?\\
\grayrow \multicolumn{2}{c}{\bf The sharp crack of shattering glass}\\
\midrule
\textbf{Flamingo} & Someone threw a glass bottle at the ground and it broke.\\
\midrule
\textbf{Qwen2} & There was a sudden loud noise, followed by the sound of glass breaking.\\
\midrule
\textbf{Ours} & I heard that too! Something smashed nearby. \\
\bottomrule
\end{tabular}
}
\caption{A human live chat example showing the different responses from models based on audio input. For comparison, we use Audio Flamingo-2~\cite{ghosh2025audio} and Qwen2-Audio-7B-Instruct~\cite{chu2024qwen2}.}
\label{tab:audioconv}  
\vspace{-7pt}
\end{table}

\end{document}